\documentclass[journal]{IEEEtran}
\usepackage{amsmath,amsfonts}
\usepackage{algorithmic}
\usepackage{algorithm}
\usepackage{array}
\usepackage{textcomp}
\usepackage{stfloats}
\usepackage{url}
\usepackage{verbatim}
\usepackage{graphicx}
\usepackage{cite}
\usepackage{makecell}
\usepackage{subfigure}

\usepackage{color}
\usepackage{diagbox}
\usepackage{float}
\usepackage{multirow}

\usepackage{xcolor}
\usepackage{listings}

\usepackage{pythonhighlight}

\begin{document}

\title{APP-Net: Auxiliary-point-based Push and Pull Operations for Efficient Point Cloud Classification}

\author{Tao Lu, Chunxu Liu, Youxin Chen, Gangshan Wu~\IEEEmembership{Member,~IEEE}, Limin Wang~\IEEEmembership{Member,~IEEE}
        % <-this % stops a space
\thanks{T. Lu, C. Liu, G. Wu, L. Wang are with the State Key Laboratory for Novel Software Technology, Nanjing  University, Nanjing, 210023, China.}% <-this % stops a space
\thanks{Y. Chen is with the Samsung Electronics (China) R$\&$D Centre, Nanjing, 210012, China.}
}

% The paper headers
\markboth{Submitted to IEEE Transactions on Image Processing}%
{Shell \MakeLowercase{\textit{et al.}}: A Sample Article Using IEEEtran.cls for IEEE Journals}

\maketitle

\begin{abstract}
Aggregating neighbor features is essential for point cloud classification. In the existing work, each point in the cloud may inevitably be selected as the neighbors of multiple aggregation centers, as all centers will gather neighbor features from the whole point cloud independently. Thus each point has to participate in the calculation repeatedly and generates redundant duplicates in the memory, leading to intensive computation costs and memory consumption. Meanwhile, to pursue higher accuracy, previous methods often rely on a complex local aggregator to extract fine geometric representation, which further slows down the classification pipeline. To address these issues, we propose a new local aggregator of linear complexity for point cloud classification, coined as APP. Specifically, we introduce an auxiliary container as an anchor to exchange features between the source point and the aggregating center. Each source point pushes its feature to only one auxiliary container, and each center point pulls features from only one auxiliary container. This avoids the re-computation issue of each source point. To facilitate the learning of the local structure of cloud point, we use an online normal estimation module to provide the explainable geometric information to enhance our APP modeling capability. Our built network is more efficient than all the previous baselines with a clear margin while still consuming a lower memory. Experiments on both synthetic and real datasets demonstrate that APP-Net reaches comparable accuracies to other networks. It can process more than 10,000 samples per second with less than 10GB of memory on a single GPU. We will release the code in https://github.com/MCG-NJU/APP-Net.

\end{abstract}


\section{Introduction}

With the growing demand for 3D applications, how to classify 3D objects with neural networks has become an important topic in recent years. Extensive work has been devoted to obtaining a higher accuracy for this task. Based on the data type and the employed networks, existing methods can be grouped into two categories. The first one is the multi-view-based 2D solution~\cite{mvcnn,mvc1,mvc2} which projects the 3D object into 2D image planes from multiple views and then applies the well-designed 2D convolutional neural network~\cite{alexnet,resnet,vgg} to learn cross-view consistent representations. These methods focus on the selection of informative views and cooperation across different views. The second solution directly learns from the 3D data with point-based networks~\cite{PointNet,PointNet++}. They focus on how to integrate the spatial relation into feature aggregation process. Several kinds of delicate local aggregators, like the point-wise MLP style and the position adaptive weighting style, are proposed to extract the fine geometric structure.

Thanks to the previous efforts, more and more works hit an accuracy of over 93\% in the most popular classification dataset of ModelNet40~\cite{wu20153d}, in the last three years. In fact, the benchmark performance has shown saturation for a long time. The detailed analysis in CloserLook3D~\cite{closerlook3d} points out that, under fair comparison, the performance gap among different local aggregators can be bridged by unified network architecture and fair training process. This conclusion reminds us to think about whether it is necessary to simply pursue a higher accuracy when designing point cloud network. Instead, we argue that, in practice, running speed and memory consumption are also important factors that should be taken into account. 

\begin{figure*}[t]
\begin{center}

\subfigure[Accuracy-Speed]{
% \label{fig:speedcomparison}
\includegraphics[width=0.3\linewidth]{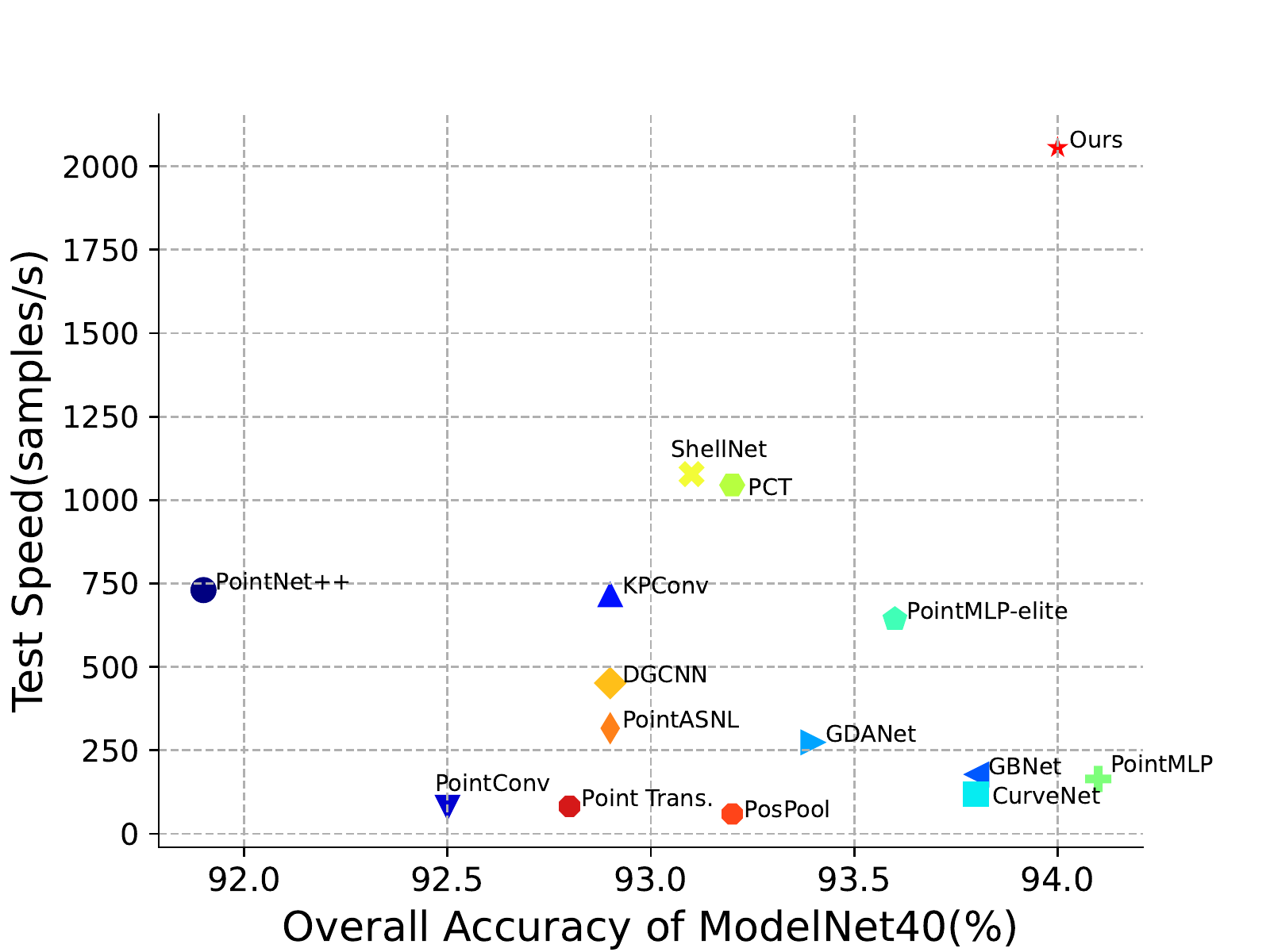}}
\subfigure[(Batch Size)-Memory]{
% \label{fig:memorycomparison}
\includegraphics[width=0.3\linewidth]{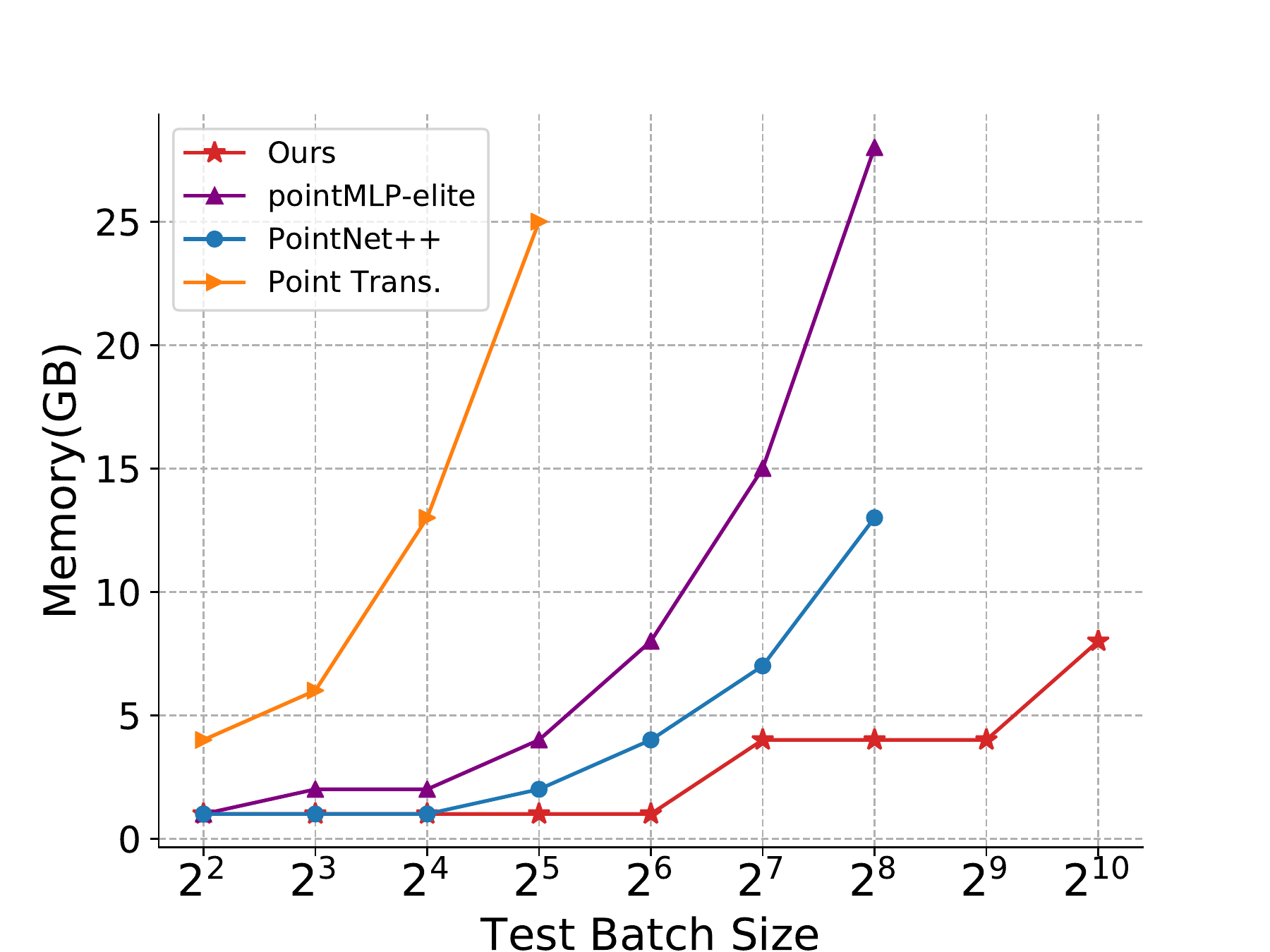}}
\subfigure[(Batch Size)-Speed]{
% \label{fig:speedbs}
\includegraphics[width=0.3\linewidth]{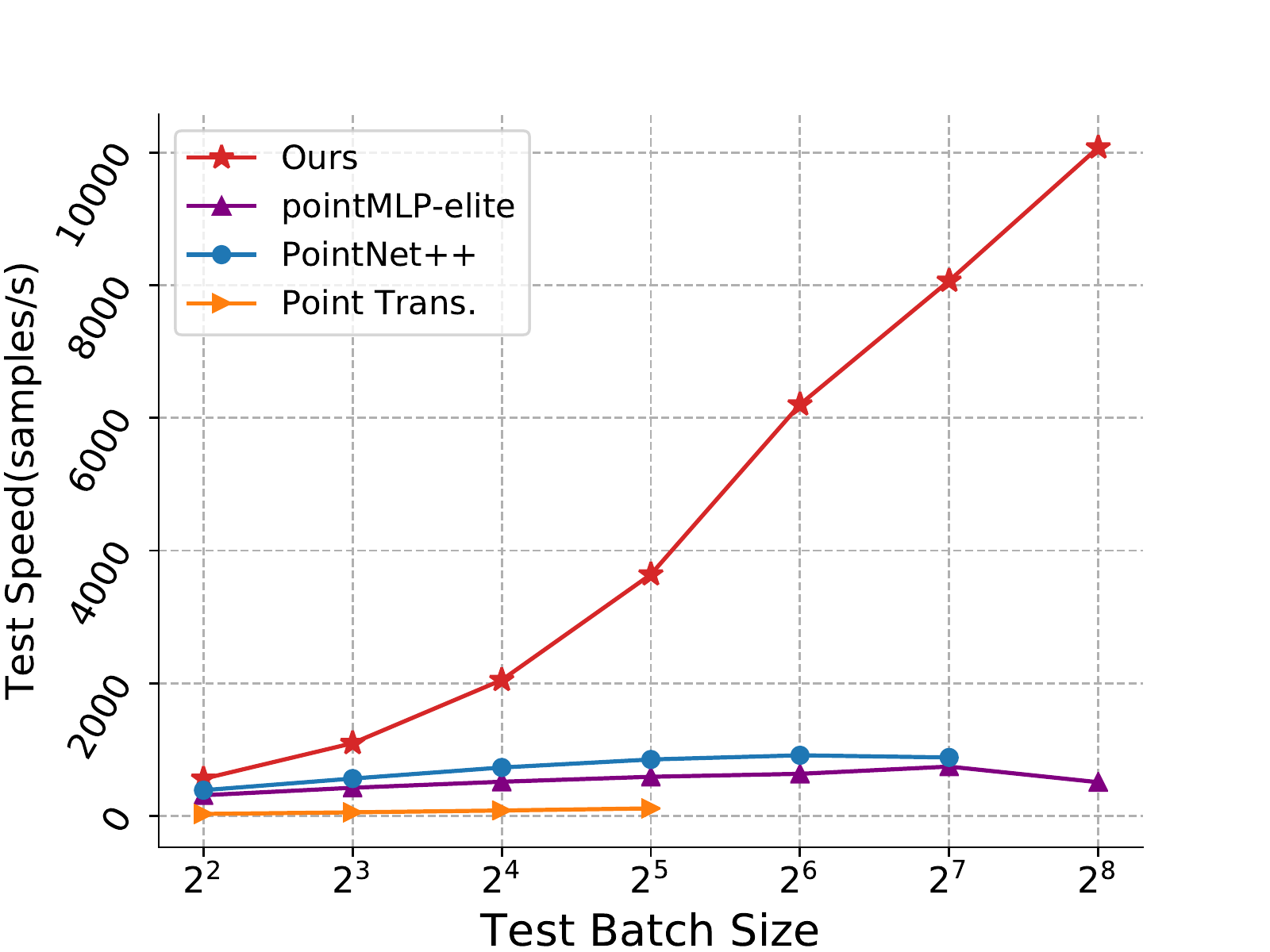}}
\end{center}
\vspace{-0.2cm}
   \caption{Quantitative Comparisons. (a) shows the accuracy-speed tradeoff on ModelNet40. We achieve the highest efficiency while maintaining comparable accuracy. (b) and (c) is the GPU memory consumption and inference speed under different inference batch sizes, respectively. In both of them, we outperform other methods with a large margin. All the experiments in (b) and (c) use 1024 points, (a) uses the default number of points in corresponding papers with a batch size=16.}
   \label{fig:overhead}
   %\label{fig:qualitativecomparison}
\end{figure*}

The essential factors responsible for the total overhead are the amount of computation and the degree of parallelism. Specifically, in the multi-view based methods, the computation and memory consumption are both linearly growing with the number of the views due to each view is processed independently. Such solutions may take several times intensive overhead to obtain slight improvement by introducing more views. For the point-based methods, due to the lack of neighbor index in the irregular point cloud, each point takes extra efforts to query and gather neighbors during the learning process. As analyzed in PVCNN~\cite{pvcnn}, the pure point-based methods are slowed down heavily by the random memory access since it corrupts the parallelism heavily by the bank conflicts. So they propose to gather neighbor features efficiently with a voxel branch by benefiting from the memory locality.

\begin{figure}[t]
\begin{center}
\label{fig:voxeltime}
\includegraphics[width=0.9\linewidth]{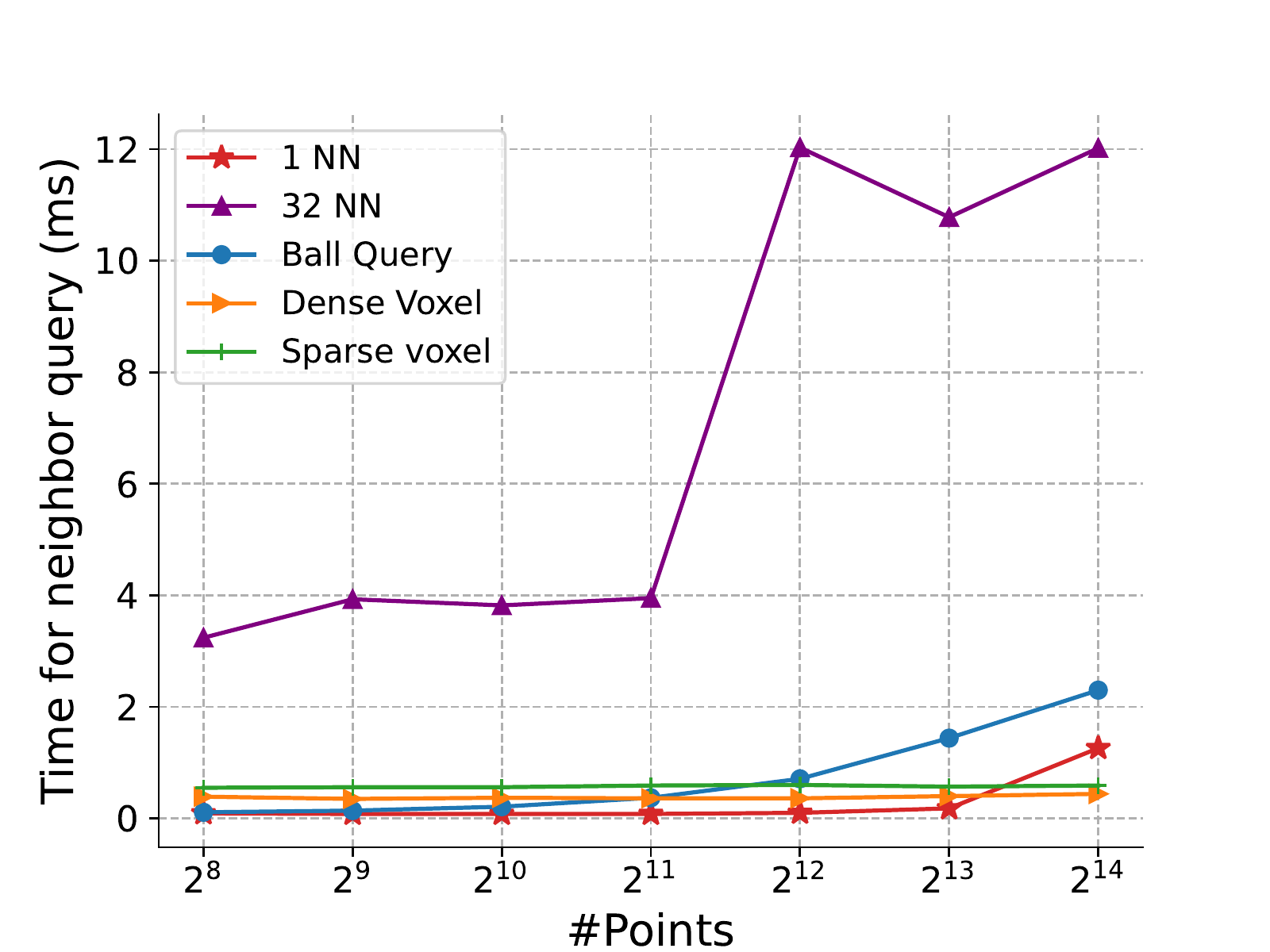}
\end{center}
\vspace{-0.3cm}
   \caption{Duration for Neighbor Query. Thanks to the parallelism, the 1-NN algorithm is efficient enough to process small scale point cloud.}
% \vspace{-0.9cm}
% \label{fig:pointnetwork}
\end{figure}

In this paper, we propose a computation and memory-efficient point-based solution to 3D classification based on the following three observations: (1) If a point is queried as multiple points' neighbors, it has to participate in computations repeatedly and occupies serveral times footprints in memory, which leads to redundant computations and memory consumption. (2) Previous architectures, except for the PointNet~\cite{PointNet}, are all designed to accomplish feature aggregation and receptive field expansion simultaneously through the overlapped neighboring area for different center points. The points in the overlapped area are inevitably queried more than one time. (3) Due to the natural sparsity in the point cloud, the extra effort on the neighbor query is unavoidable. Even for the voxel-aided neighbor gathering, the dense voxelization manner still wastes extra resources on a large amount of blank voxels. The sparse manner also suffers from memory bank conflicts in the scatter and gather process. According to Figure~\ref{fig:voxeltime}, although the kNN algorithm costs too much time, its 1-NN variant shows great efficiency surpassing all the other methods. Based on the first two observations, we conclude that one key towards an efficient point-based classification network is {\em to decouple the process of feature aggregation and receptive field expansion}. In addition,  the third observation suggests that for the classification task (often uses 1024$\sim$4096 points), the 1-NN algorithm is efficient enough to query neighbors. 

 Based on the above analysis, we propose a new network whose overall computation is reduced to linear complexity, and it only costs linear memory occupations. Specifically, we aggregate features and expand the receptive field in separate steps. During aggregation, to avoid repeated operations to each point, we introduce an auxiliary container as a proxy to exchange features among points. The container consists of a series of pre-specified anchors in 3D space. Then each point is processed by two operations: first, push its feature to only one nearest auxiliary anchor; second, pull the feature from only one nearest auxiliary anchor. According to the corresponding anchor, the point cloud is split into several non-overlapped blocks. Points closest to the same auxiliary anchor in Euclidean space will fall into the same block and accomplish the features exchange automatically. Each anchor only costs a tiny maintenance overhead. To avoid the artifact introduced by the auxiliary anchor, we propose a novel push and pull pair through which the influence from the anchor is reducible. To enable receptive field expansion, we introduce a second auxiliary container to produce a different partition of the whole point cloud. Combining the two-stage block partitions, we obtain an equivalent receptive field expansion. Finally, to facilitate learning the local structure in the early layer, we use an online normal estimation module to provide explainable geometric information to enhance our APP block's modeling capability. 
 
\begin{figure*}[t]
\begin{center}
\subfigure[PointNet.]{
% \label{fig:pointnet}
\includegraphics[width=0.4\linewidth]{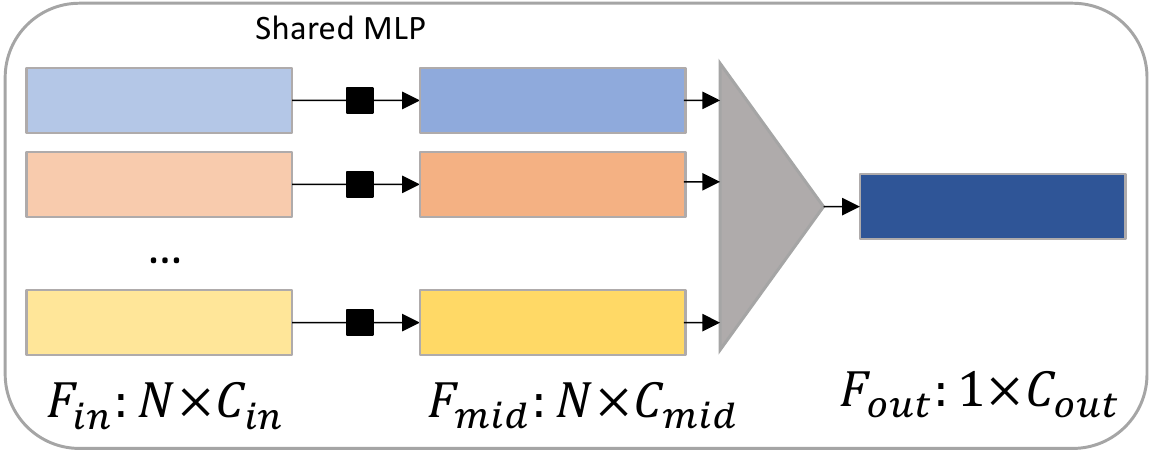}}
\subfigure[PointNet++.]{
% \label{fig:pointbased}
\includegraphics[width=0.4\linewidth]{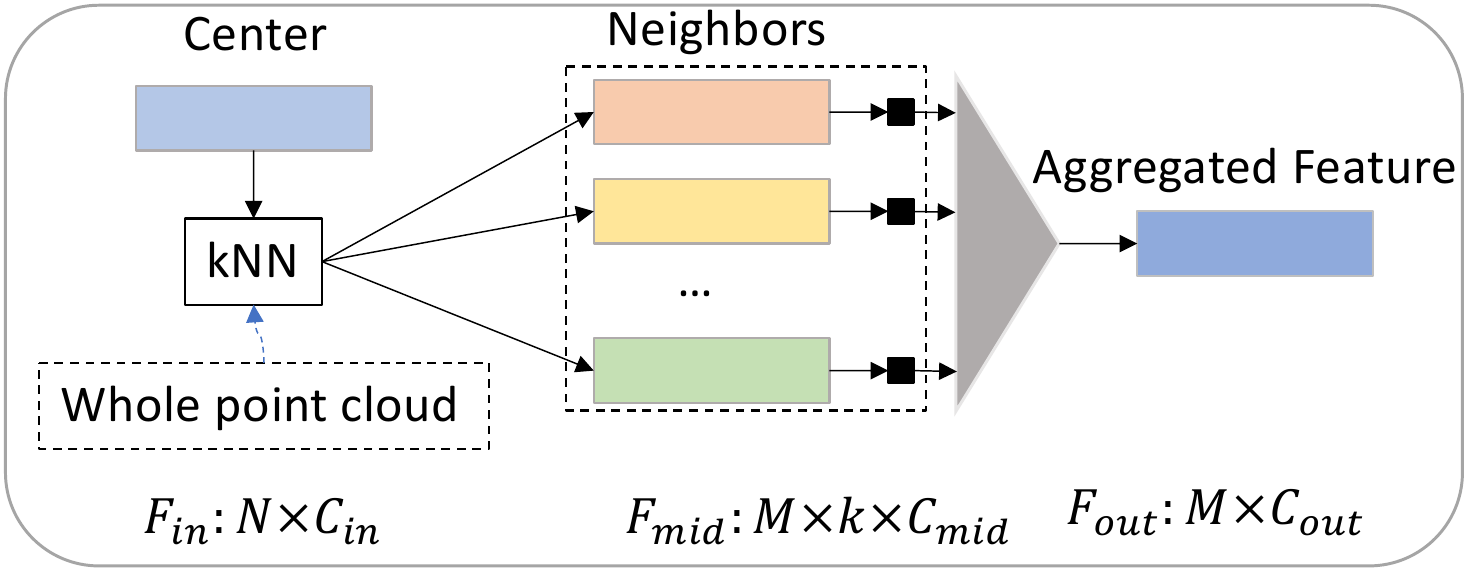}}
\subfigure[Point Cloud Transformer.]{
% \label{fig:pointbased}
\includegraphics[width=0.4\linewidth]{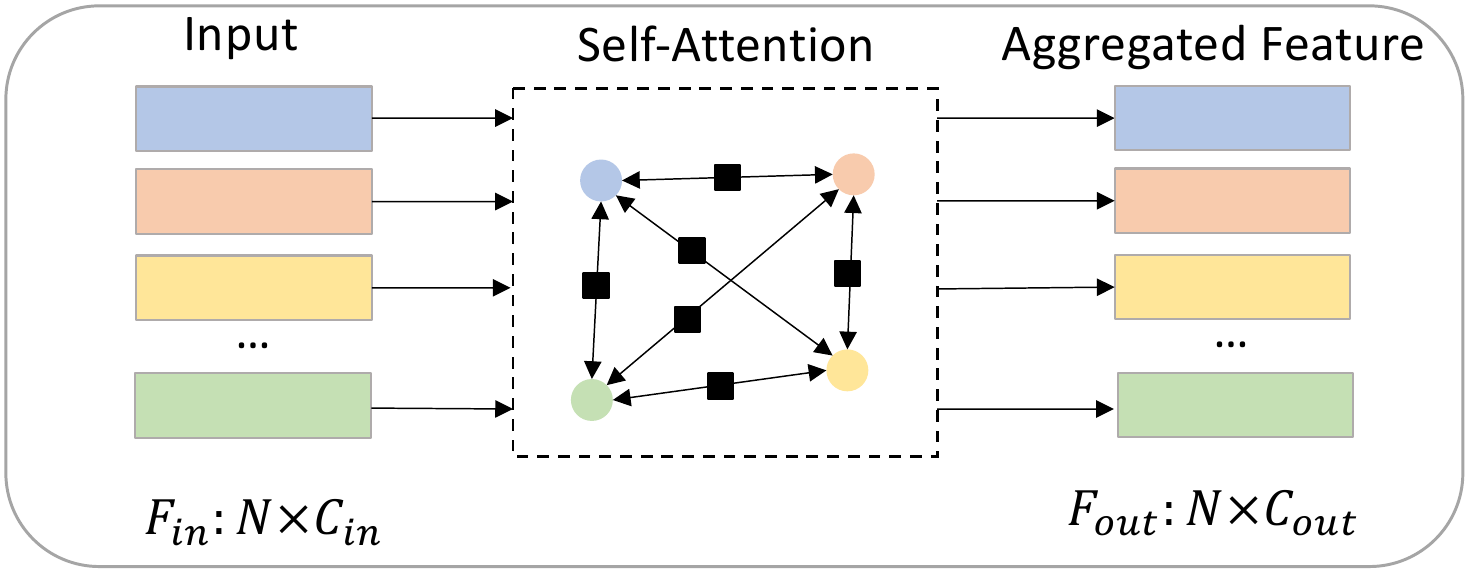}}
\subfigure[APP-Net.]{
% \label{fig:ours}
\includegraphics[width=0.4\linewidth]{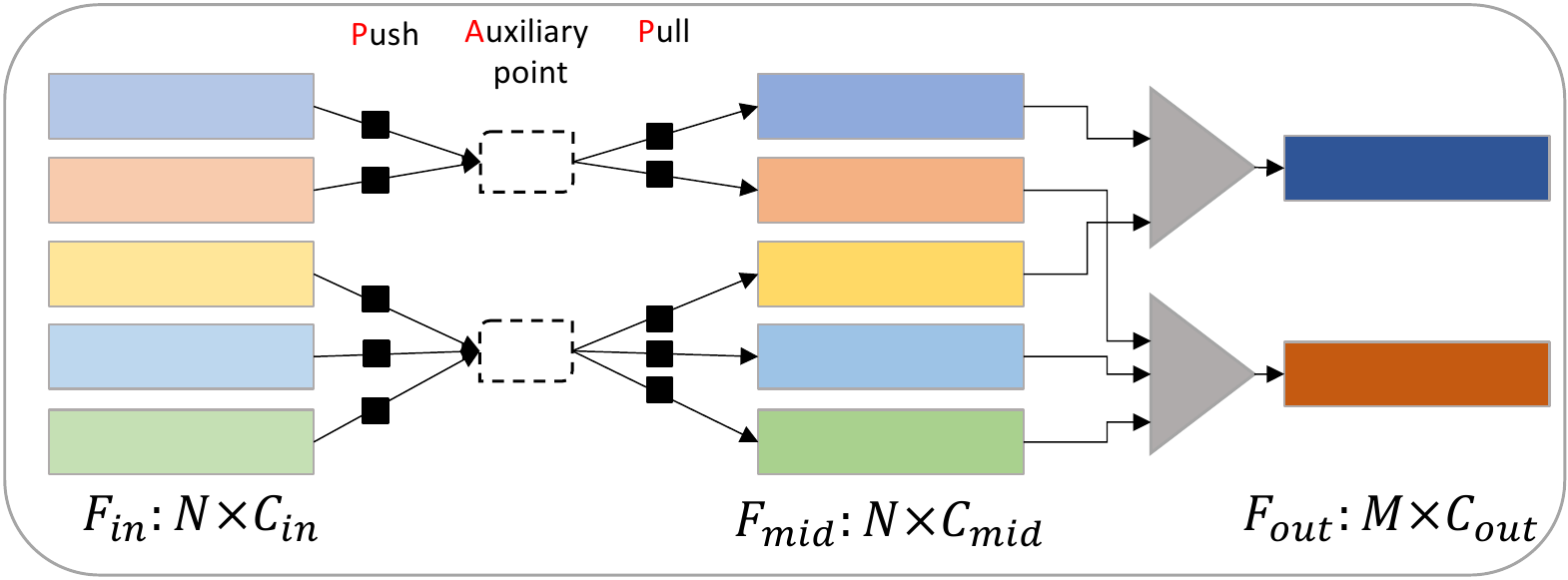}}
\end{center}
\vspace{-0.3cm}
   \caption{Comparisons among Point Cloud Networks, including the PointNet, PointNet++, Point Cloud Transformer, and our proposed APP-Net. The \textcolor[RGB]{0,0,0}{\bf black boxes} in each method refer to the learnable parts.}
% \vspace{-0.9cm}
\label{fig:pointnetwork}
\end{figure*}

 The auxiliary-anchor-based push and pull operations pair, the so-called APP operator, achieve a huge reduction in the memory consumption and computation complexity while maintaining comparable accuracy to the previous methods. The comparisons among different styles of network structure, including PointNet~\cite{PointNet}, PointNet++~\cite{PointNet++}, point cloud transformer~\cite{guo2021pct}, and the proposed APP-based network, are depicted in Fig~\ref{fig:pointnetwork}. It is easy to see that the overhead of the APP is linear to the number of input points. We conducted a detailed quantitative analysis of the running speed and memory consumption. As depicted in Fig~\ref{fig:overhead}, we achieve an inference speed of more than 2000 samples/second with batch size=16. Among those networks outperforming 92\%, we are clearly faster than the second efficient network, ShellNet~\cite{shellnet}. Moreover, our network consumes a remarkably low GPU memory. According to Fig~\ref{fig:overhead}, APP-Net only costs memories less than 10GB with a batch size=1024. Correspondingly, our machine's maximum supported batch size for Point Transformer~\cite{zhao2021point} is 64, which costs more than 30GB of memory. And the lightweight version of PointMLP~\cite{ma2022rethinking} consumes more than 25GB with a maximum batch size of 256. Furthermore, according to Fig~\ref{fig:overhead}, we can even achieve a speed more than {\bf10,000 samples/s} with a batch size of 256, which is {$\bf5\times$} faster than the peak of other baselines. More details and analysis are presented in the following sections. In summary, the main contributions of this paper are:
 
 \begin{enumerate}
     \item We propose to decouple the feature aggregation and receptive field expansion process to facilitate redundancy reduction.
     \item We propose an auxiliary-anchor-based operator to exchange features among neighbor points with linear computation complexity and linear memory consumption.
    %  \item We design a reducible style to eliminate the influence of the intermediate process.
     \item We propose to use the online normal estimation to improve the classification task.
     \item Experiments show that the proposed network achieves remarkable efficiency and low memory consumption while keeping competitive accuracy.
 \end{enumerate}

\section{Related Work}
In this section, we review some remarkable works in 3D object classification. According to the data type, we divide those methods into multi-view-based and point-cloud-based methods.

\subsection{Multi-view Based Methods}
%\vspace{-0.2cm}
Multi-view-based methods consider learning point features with the mature CNN by projecting the 3D object into a series of 2D images from multiple views. Some works~\cite{mvcnn,mvc1,yu2018multi,yang2019learning,sun2020drcnn,feng2018gvcnn,chen2018veram} devoted to investigating how to fuse the features with pooling policy. Yang et al.~\cite{mvc3} propose to combine with the guidance of the inter-relationship among views. MVTN~\cite{hamdi2021mvtn} learns to search a series of more informative viewpoints. Wei et al.~\cite{mvc2} treats the different views as nodes in a graph, which facilitates the use of GCN to exploit the relation among views. 
Carlos et al.~\cite{esteves2019equivariant} proposes to use rotation equivariant convolution on multi-views.Liu et al.~\cite{DBLP:journals/tip/LiuHLZ21} introduces a more complex setting of fine-grained classification and proposes to explicitly detect the regions of interest. 

In general, the projection process is view-dependent. It requires processing many views to alleviate the geometrical information loss caused by the projection, making it intensive to analyze each 3D object. And it's challenging to fuse the view-wise features from different views to obtain a consistent and discriminative representation.
%\vspace{-0.05cm}

\subsection{Point Based Methods}
%\vspace{-0.2cm}
This is a widely used category in 3D classification. Motivated by the 2D CNN~\cite{krizhevsky2012imagenet,long2015fully}, many works are designed to aggregate local point features. The local area is determined by the distance in Euclidean space or the topology. DGCNN~\cite{DGCNN} constructs a graph to learn with the connections between nodes. PointNet++~\cite{PointNet++} provides a standard paradigm for fully point-based method. It points out how to locate the neighbor area and aggregate local features point-wisely. Subsequent works~\cite{shellnet,PointASNL} improve the designs of local area structure, downsampling methods, and local aggregator. Different from PointNet++~\cite{PointNet++}, some works design convolution-like operators for point cloud. SpiderNet~\cite{SpiderNet} uses Taylor expansion to approximate the spatial distribution of filters. KPConv~\cite{KPConv} specifies a series of kernel points to implement convolution. PointConv~\cite{pointconv} directly learns the values of the filter through coordinates. And PAConv~\cite{xu2021paconv} proposes to assemble the pre-specified weight banks according to the spatial relation. The recent hot topic, transformer, has started to show its power in the point cloud. PCT~\cite{guo2021pct} is the first totally transformer-based network which conducts a global self-attention in each layer. It cuts down the process to query neighbors because each point serves as all the other points' neighbors. Point Transformer~\cite{zhao2021point} enhances the local aggregator with a Transformer-like operator.  Point2SpatialCapsule~\cite{wen2020point2spatialcapsule} proposes to not only model the geometric information but also model the spatial relation with the Capsule Network~\cite{sabour2017dynamic}. L3DOC~\cite{liu2021l3doc} introduces lifelong learning to extending the 3D classification task into open environments. DSDAN~\cite{9483674} investigates the problem of cross-domain 3D classification. These methods have achieved remarkable accuracy. However, efficiency and low memory consumption are not their main targets. Some of the few attempts for efficiency explore by eliminating the existing architectures. ShellNet~\cite{shellnet} proposes to shrink some heavy operations the PointNet++ to construct a lightweight architecture, PointMLP-Elite leverages the bottleneck architecture to reduce the feature transforming burden. Although they have shown effectiveness in accelerating, they do not solve the problem of redundant resource calls. So they leave us with a huge room for eliminating the overhead. We achieve the most efficient network for point cloud classification with the proposed APP operator.

\section{Method}

\subsection{Background and Insights}
\label{sec:analysis}

\subsubsection{Preliminaries} The source point cloud with $N$ points is denoted as $\mathbf{P}=\{\mathbf{p}_1, \mathbf{p}_2, ..., \mathbf{p}_N\}$. The target is to aggregate current source features into a set of center points $\mathbf{Q}=\{\mathbf{q}_1, \mathbf{q}_2, ..., \mathbf{q}_M\}$. The general process of a local point feature aggregation is as follows:

\begin{equation}
\label{eq:conv}
\mathbf{g}_i = R(\{G(r(\mathbf{q}_{i}, \mathbf{p}_{j}, [\mathbf{f}_{i}, \mathbf{f}_{j}]), \mathbf{f}_j)|\mathbf{p}_{j}\in N(\mathbf{q}_i)\})
\end{equation}

\noindent
where $\mathbf{q}_i$ is the center point, $\mathbf{g}_i$ is the output feature for the center point. $N(\mathbf{q}_i)$ queries neighbors for $\mathbf{q}_i$. $\mathbf{f}_{*}\in\mathbb{R}^{C_{in}}$ denotes the input feature of $\mathbf{p}_{*}$ or $\mathbf{q}_{*}$. $r(\mathbf{q}_{i}, \mathbf{p}_{j}, [\mathbf{f}_{i}, \mathbf{f}_{j}])$ measures the relation between the neighbor point and the center point. Most methods mainly consider the position relation, and others combine the feature relation. $G$ and $R$ refer to the features transformation and reduction, respectively. The reduction function is MAX Pooling in most cases.

\subsubsection{Analysis and Insights} For the common point-based architecture~\cite{PointNet++}, each center point selects neighbors from the source points independently. Each source point would be selected as multiple centers' neighbors to accomplish the receptive field expansion. Thus, each source point has to repeatedly participate in the aggregation process and occupies several duplicates of memories to facilitate parallelization. The advantage of this manner is obvious: it enables the neighbor point to adaptively contribute to different center points according to the current spatial and feature relation, and it combines the feature extraction and receptive field expansion in one step. However, it inevitably induces extra computations and memory consumption for redundant operations. In the first layer of PointNet++~\cite{PointNet++}, $N=1024$, $M=512$, each center point queries 32 neighbors, which indicates that each source point would be replicated and re-computed for $\frac{512\times32}{1024}=$16 times.

\begin{figure*}[tbp]
\begin{center}
\includegraphics[width=0.9\linewidth]{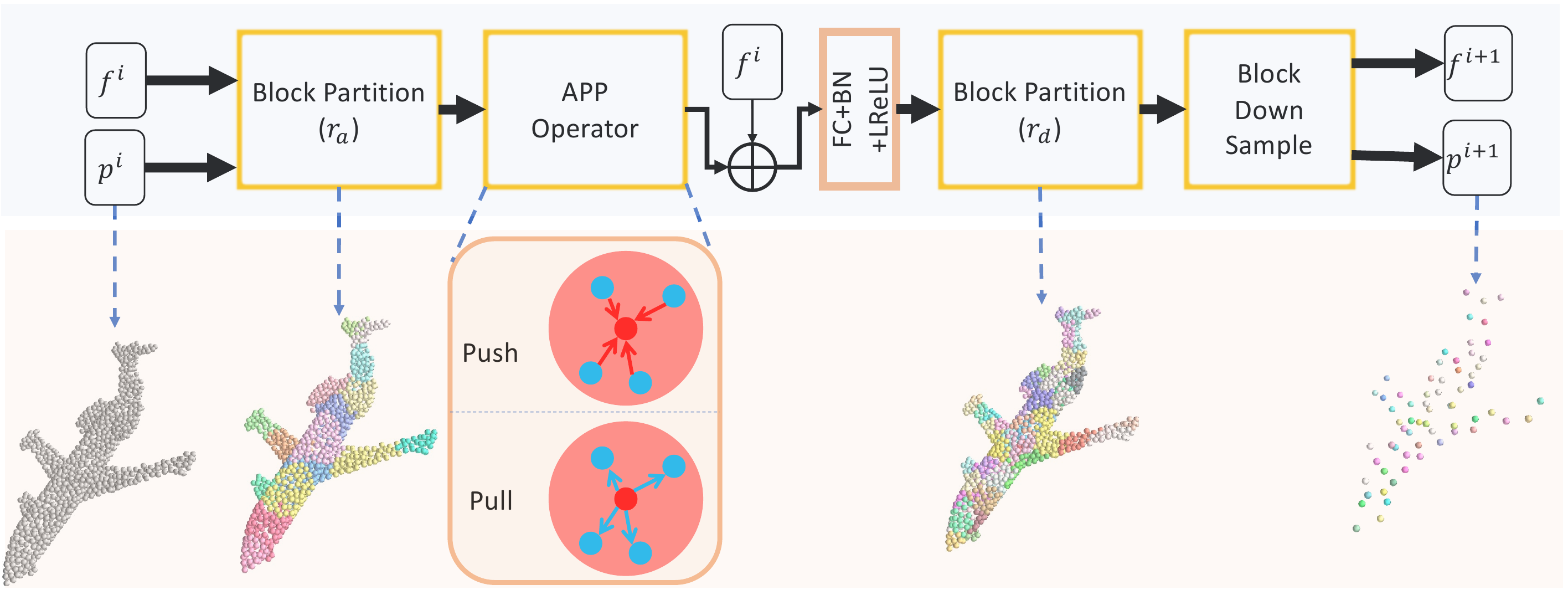}
\end{center}
   \caption{The APP Block Structure. The inputs are coordinates with features, and the output is downsampled coordinates and updated features.}
\end{figure*}

To avoid using redundant resources, we notice that all the previous methods conduct the aggregation procedure from the view of the center point $\mathbf{Q}$, and each center point is agnostic of how many times other center points have accessed the neighbor. To limit the access times of each source point, we turn to guide the aggregation from the view of the source point. In this paper, we proposed a so-called APP (\textcolor[rgb]{1,0,0}{a}uxiliary-based \textcolor[rgb]{1,0,0}{p}ush and \textcolor[rgb]{1,0,0}{p}ull) operator which introduces the auxiliary point set $\mathbf{A}$ as a bridge between the source points and the center points. Each source point $\mathbf{p}$ only pushes its information to one auxiliary point, and each center point $\mathbf{q}$ only pulls information from one auxiliary point, i.e., during the whole feature aggregation process, each source point only participates in emitting feature for one time. And the center point only participates in gathering features for one time. Such paired operations linearize the complexity of computation. Although reducing some overhead, auxiliary points may also introduce incorrect dependencies on the artifacts. To eliminate the influence of auxiliary points, we implement the operation pair in a novel manner, with which the influence of auxiliary points is reducible. The core idea is to find a group of functions \{$\alpha$, $\beta$, $\gamma$, $\epsilon$\} which satisfy the following relation:
\begin{equation}
\label{eq:opcondition}
    \gamma(x\to y)=\epsilon[\alpha(x\to a), \beta(a\to y)]
\end{equation}
\noindent
where $\alpha(x\to a)$ and $\beta(a\to y)$ denote the process of pushing features from $x$ to the auxiliary point $a$ and pulling features from $a$ to destination point $y$ respectively. $\epsilon$  combines $\alpha(*)$ and $\beta(*)$ in a reducible way such that the resulted $\gamma(x\to y)$ only depends on the $x$ and $y$, not affected by the auxiliary point. In the following sections, we will introduce some instantiations of the proposed \{$\alpha$, $\beta$, $\gamma$, $\epsilon$\}.

\subsection{Auxiliary-based Push and Pull Operations}
\subsubsection{Auxiliary Point Generation}
\label{sec:blockpart}
As introduced above, the auxiliary points serve as bridges to pass information among local points. For simplicity, we directly downsample the original point cloud to obtain a subset $\mathbf{A}=\{\mathbf{a}_0, \mathbf{a}_1,...,\mathbf{a}_{\frac{N}{r_a}}\}$, where the $r_a$ is a downsample ratio. Following RandLA-Net~\cite{randla}, we adopt the Random Sample instead of the widely-used FPS (Farthest Point Sample)~\cite{PointNet++} for further acceleration. Although the uniformity is slightly disturbed, the experimental results show 
little sensitivity to the sampling strategy. As mentioned in Section~\ref{sec:analysis} that each source point only emits its information to one auxiliary point, we conduct a 1-NN query for every source point from the auxiliary point set $\mathbf{A}$. For a point $\mathbf{p}_i$, we denote its auxiliary point as $A(\mathbf{p}_i)$. Points close to each other in Euclidean space naturally choose the same auxiliary point. As a result, the whole point cloud is partitioned into several non-overlapped blocks $\{B(\mathbf{a}_0), B(\mathbf{a}_1), ..., B(\mathbf{a}_{\frac{N}{r_a}})\}$, where $B(\mathbf{a}_i)$ denotes the source point set whose 1-NN auxiliary point is $\mathbf{a}_i$.

\subsubsection{The Design of Operation Groups}
According to the previous analysis, a reducible operation group conforming to Equation~\ref{eq:opcondition} eliminates the auxiliary point's influence to prevent artifacts. Given two points $x$ and $y$ belonging to the same block, we first instantiate the $\epsilon(*)$ with two optional basic operators: element-wise multiplication and element-wise addition. The Equation~\ref{eq:opcondition} is rewritten as follows,

\begin{equation}
\label{eq:multiplication}
    \gamma(x\to y)=\alpha(x\to a)\cdot \beta(a\to y),
\end{equation}

\noindent
or

\begin{equation}
\label{eq:addition}
    \gamma(x\to y)=\alpha(x\to a) + \beta(a\to y).
\end{equation}

\noindent
For simplicity, we will use $\alpha$ to denote $\alpha(x\to a)$ and $\beta$ to denote $\beta(a\to y)$ in the following. Then we construct satisfying $\alpha$ and $\beta$ for the two operators. For the addition operator, it's obviously that if $\alpha$ and $\beta$ are linear mappings, i.e.

\begin{small}
\begin{equation}
\begin{aligned}
\gamma(x-y)&=W\times{(x-y)},\\&=\alpha + \beta
, where\left\{
\begin{aligned}
%\nonumber
\label{eq:addbased}
    \alpha &= W\times (x-a),\\
    \beta &= W\times (a-y),\\
\end{aligned}
\right.
\end{aligned}
\end{equation}
\end{small}

\noindent
where $W$ is a weight matrix shared by $\alpha$ and $\beta$, the influence from the auxiliary point is easily eliminated, and the resulted $\gamma$ is also a linear mapping function. Following a similar idea, we construct the $\alpha$ and $\beta$ with exponential mapping for the multiplication operator. Specifically, 

\begin{equation}
\begin{aligned}
\gamma(x-y)&=e^{W\times(x-y)},\\&=\alpha\cdot\beta
, where\left\{
\begin{aligned}
%\nonumber
\label{eq:multibased}
    \alpha &= e^{W\times(x-a)},\\
    \beta &= e^{W\times(a-y)}.\\
\end{aligned}
\right.
\end{aligned}
\end{equation}
The above designs are based on a single operator (multiplication or addition). When we jointly employ multiple basic operators for $\eta$, more sophisticated operation groups can be derived. According to the "sum-difference-product" formula for trigonometric functions, we obtain $sin$ and $cos$ based operator groups as follows,

\begin{small}
\begin{equation}
\begin{aligned}
\gamma(x-y) &=cos(W\times(x-y))\\&=\alpha[0]\beta[0]-\alpha[1]\beta[1]
, where\left\{
\begin{aligned}
%\nonumber
    \alpha[0] &= cos(W\times(x-a))\\
    \alpha[1] &= sin(W\times(x-a))\\
    \beta[0] &= cos(W\times(a-y))\\
    \beta[1] &= sin(W\times(a-y))\\
\end{aligned}
\right.,
\end{aligned}
\end{equation}
\end{small}

\begin{small}
\begin{equation}
\begin{aligned}
\gamma(x-y) &=sin(W\times(x-y)),\\&=\alpha[0]\beta[1]+\alpha[1]\beta[0]
, where\left\{
\begin{aligned}
%\nonumber
    \alpha[0] &= sin(W\times(x-a)),\\
    \alpha[1] &= cos(W\times(x-a)),\\
    \beta[0] &= sin(W\times(a-y)),\\
    \beta[1] &= cos(W\times(a-y)).\\
\end{aligned}
\right.
\end{aligned}
\end{equation}
\end{small}

We believe that there are infinite operation groups satisfying the reducible philosophy. And we have no intention of exhausting those potential superior combinations. We adopt the above operation groups to form the basis of this work. They decompose the inter-communication process within the local area. For a local patch with $n$ points, to pass information between every point-pair, previous methods like~\cite{zhao2019pointweb} induce $\mathcal{O}(n^2)$ overhead. While with the proposed operation groups, all points share the same $\alpha(x-a)$ to obtain the information of x. The number of all possible $\alpha(*)$ is $n$. Thus we obtain a $\mathcal{O}(n)$ complexity overhead. Furthermore, due to some mathematical properties (parity, reciprocal relation, opposite relation, and so on) of the basic operators, some parts of the operations can be reused by other parts, which also contributes a lot to lowering the consumption of computation and memory resources. We summarize the reusable parts as follows,

\begin{itemize}
    % \item {\bf Addition}-based: $\beta=-\alpha$;
    \item {\bf Exponential}-based: $\beta=\frac{1}{\alpha}$;
    \item {\bf Cosine}-based: $\beta[0]=\alpha[0]$, $\beta[1]=-\alpha[1]$;
    \item {\bf Sine}-based: $\beta[0]=-\alpha[0]$, $\beta[1]=\alpha[1]$.
\end{itemize}
% 现在我们构建了4种基础操作符

\subsection{Push and Pull based Feature Aggregation}

One key advantage of the point cloud is the preservation of geometry. Previous local aggregators mine the 3D structure information through modeling the spatial relation among points. According to how to mine the spatial relation, they are classified into different categories, like the point-wise MLP-based or adaptive-weight-based methods. For the point-wise MLP type, the Equation~\ref{eq:conv} is usually instantiated as follows,

\begin{equation}
\label{eq:pointwise}
\mathbf{g}_i = R(\{MLP(\mathbf{q}_{i}-\mathbf{p}_{j}, \mathbf{f}_j)|\mathbf{p}_{j}\in N(\mathbf{q}_i)\}).
\end{equation}

\noindent
The adaptive-weight-based methods generate position adaptive kernel to weight neighbors. The Equation~\ref{eq:conv} is instantiated as follows,

\begin{equation}
\label{eq:adaptive}
\mathbf{g}_i = R(\{\mathcal{W}(\mathbf{q}_{i}-\mathbf{p}_{j})\cdot\mathbf{f}_j|\mathbf{p}_{j}\in N(\mathbf{q}_i)\})
\end{equation}

 \noindent
 where $\mathcal{W}(*)$ generates the kernel weight according to the spatial relation.
 
In this part, we show examples of how to build the two types local aggregators based on the above operation groups. Only parts of the proposed operation groups will be exhibited, and the others can be implemented by following similar steps or referring to our released code. The process consists of Position Encoding, Push-step, Pull-step, Channel Mixing, and Block-Based Down Sample.

\noindent
{\bf -Position Encoding} At the beginning of every setting, we implement a simple and fast position encoding to lift the original coordinate from a 3-channel vector to an embedding of $C_{in}$ channels. The global sharing encoding function $\phi(*)$ is implemented by [FC Layer + BatchNorm + Leaky ReLU]. We use the position encoding, instead of the original coordinate, to cope with the features in all the following steps.

\noindent
{\bf -Push \& Pull} In the Push-step, the information of the source point is delivered to the auxiliary point. And in the Pull-step, each point gathers features from the corresponding auxiliary point. We introduce how to build a point-wise MLP-based and adaptive-weight-based local aggregator as examples.

\begin{enumerate}
    \item {\bf Point-wise MLP} According to Equation ~\ref{eq:pointwise}, a point-wise MLP style aggregator requires concatenating features and spatial relation. Thus the push step is defined by

    \begin{equation}
        \mathbf{g}_{\mathbf{p}_i\to A(\mathbf{p}_i)}=W\times[\mathbf{f}_i, \phi(\mathbf{p}_i)-\phi(A(\mathbf{p}_i))]\label{eq:push},
    \end{equation}

    \noindent
    where the $W\in\mathbb{R}^{2C_{in}\times C_{in}}$ is a weight matrix corresponding to the one in Equation~\ref{eq:addbased}. This step pushes the source point's feature to the auxiliary point according to the spatial relation. Mind that all the source points belonging to the same auxiliary block will push their feature to the same auxiliary point. Thus the final feature in the auxiliary point is computed by
    
    \small
    \begin{equation}
        \mathbf{g}_{A(\mathbf{p}_i)}=\frac{1}{|B(A(\mathbf{p}_i))|} \sum_{\substack{\mathbf{p}_j\in \\B(A(\mathbf{p}_i))}}{W\times[\mathbf{f}_j, \phi(\mathbf{p}_j)-\phi(A(\mathbf{p}_i))]}.
    \end{equation}
    
    \noindent
    In the pull step, following Equation~\ref{eq:addbased}, we apply an inverse operation for each source point to pull features from the corresponding auxiliary point. The inverse operation is formulated as follows:
    
    \small
    \begin{align}
        \mathbf{g}_i&=\mathbf{g}_{A(\mathbf{p}_i)\to \mathbf{p}_i}, \notag\\
        &=\mathbf{g}_{A(\mathbf{p}_i)}+ W\times[-\mathbf{f}_i, \phi(A(\mathbf{p}_i))-\phi(\mathbf{p}_i)]\notag\\
        &=\frac{1}{|B(A(\mathbf{p}_i))|}\sum_{\substack{\mathbf{p}_j\in \\B(A(\mathbf{p}_i))}}{W\times[\mathbf{f}_j-\mathbf{f}_i, \phi(\mathbf{p}_j)-\phi(\mathbf{p}_i)]}\label{eq:assign}.
    \end{align}
    
    According to the output, the resulted $\mathbf{g}_i$ is only computed by the point feature and spatial relation. The auxiliary point $A(\mathbf{p}_i)$ only serves to provide a neighbor query. Different from the common practice, we adopt AVG Pooling as the reduction function. One more difference is, for leveraging the reusable computation, Equation~\ref{eq:assign} finally concatenates the feature difference and spatial relation. To realise the concatenation of original feature and spatial relation, one can modify the $\beta$ operation by replacing the feature with zeros vector, i.e. $\beta=W\times[\mathbf{0}, \phi(A(\mathbf{p}_i))-\phi(\mathbf{p}_i)]$. And the $\beta$ operation cannot reuse the results of $\alpha$. We will discuss its influence in the experiment section.
    
    \item {\bf Adaptive Weight} Following similar steps, we can easily construct an adaptive weight aggregator. Here we use the {\bf Exponential}-based operation groups. The push step is defined by
    
    \begin{equation}
        \mathbf{g}_{\mathbf{p}_i\to A(\mathbf{p}_i)}=\mathbf{f}_i\cdot e^{W\times[\phi(\mathbf{p}_i)-\phi(A(\mathbf{p}_i))]},
    \end{equation}
    
    \noindent
    where $W\in\mathbb{R}^{C_{in}\times C_{in}}$ generates weight kernel according to the spatial relation. The $e^{(*)}$ provides channel-wise weights to the input feature. The features in $A(p_i)$ is 
    
    \begin{equation}
        \mathbf{g}_{A(\mathbf{p}_i)}=\frac{1}{|B(A(\mathbf{p}_i))|}\sum_{\substack{\mathbf{p}_j\in \\B(A(\mathbf{p}_i))}}\mathbf{f}_j\cdot e^{W\times[\phi(\mathbf{p}_j)-\phi(A(\mathbf{p}_i))]}
    \end{equation}
    
    Then the pull step is formulated as 
    
    \begin{align}
    \mathbf{g}_i&=\mathbf{g}_{A(\mathbf{p}_i)\to \mathbf{p}_i} \notag\\
    &=\mathbf{g}_{A(\mathbf{p}_i)}\cdot e^{W\times[\phi(A(\mathbf{p}_i))-\phi(\mathbf{p}_i)]}\notag \\
    &=\frac{1}{|B(A(\mathbf{p}_i))|}\sum_{\substack{\mathbf{p}_j\in \\B(A(\mathbf{p}_i))}}\mathbf{f}_j\cdot e^{W\times[\phi(\mathbf{p}_j)-\phi(\mathbf{p}_i)]}\label{eq:expassign}.
\end{align}

    \noindent
    Essentially speaking, the final output is an instantiation of Formula~\ref{eq:adaptive}, whose reduction function is AVG Pooling. It weights each channel according to the spatial relation with neighbors.
    
\end{enumerate}

\noindent
{\bf -Channel Mixing} The push and pull steps are efficient operations for mixing the features among points in the local area. However, there exist two obstacles towards a better representation: first, the use of AVG Pooling tends to obscure some high-frequency patterns in each local area. Then, some push and pull operations, like the {\bf Exponential}-based groups, only conduct channel-wise weighting without inter-channel interaction, which damages the model capacity. Thus we employ a skip connection from the input features to make up for the high-frequency information. And we introduce a Fully Connection layer to enhance channel mixing. The feature is updated by 
\begin{equation}
	\mathbf{g}_{i} = \delta([\mathbf{g}_i, \mathbf{f}_i]).
\end{equation}
where $\delta(*)$ is a non-linear function constituted by \{FC+BatchNorm+LeakyReLU\}. The resulted $\mathbf{g}_i$ is of $C_{out}$ channels.

\begin{algorithm}[t]
  \caption{Exponential-based adaptive weight aggregator}
  \begin{algorithmic}\label{alg:expaw}
  \STATE
    \begin{python}
    # points: [N, 3], F: [N, C]
    # r_a, r_d
    
    # Block Partition
    aux_points = rand_choice(points, N/r_a) 
                                # [N/r_a, 3]
    idx_PtoA = one_nn(points, aux_points) 
                                    # [N, 1]
    # push and pull
    pos_enc = Linear_BN_LReLU(points) # [N, C]
    kernel = exp(Linear(pos_enc)) # [N, C]
    F_weighted = F * kernel # [N,C]
    F_PtoA = scatter_mean(F_weighted, idx_PtoA) 
                                    # [N/r_a, C]
    F_AtoP = gather(F_PtoA, idx_PtoA) # [N, C]
    aggregated_F = F_AtoP / kernel # [N,C]
    #Channel Mixing
    new_F = Linear_BN_LReLU([F, new_F]) 
                            # [N,C_out]
    # Block Based Down Sample
    centroids = rand_choice(points, N/r_d) 
                                # [N/r_d, 3]
    idx_PtoC = one_nn(points, centroids) 
                                    # [N, 1]
    out_F = scatter_max(new_F, idx_P2C) 
                        # [N/r_d,C_out]
    
    return out_F, centroids
    \end{python}
  \end{algorithmic}
\end{algorithm}

\noindent
{\bf -Block Based Downsampling}
After processing by Push-step and Pull-step, the output point cloud still has the same number of points as the input point cloud, which is similar to the PointNet layer~\cite{PointNet}. We design a block-based down sample strategy to reduce the intermediate overhead further. Like the operation in Section~\ref{sec:blockpart}, we downsample the point cloud with a rate of $\mathbf{r}_d$ and re-split the whole cloud into serval non-overlap blocks $\{\mathbf{D}_0, \mathbf{D}_1, ..., \mathbf{D}_{\frac{N}{\mathbf{r}_d}}\}$ based on 1-NN algorithm. Then, for all the points belonging to the same block $\mathbf{D}_i$, their features are aggregated by a MAX pooling function as follows:
\begin{equation}
\mathbf{g}_{\mathbf{d}_i} = MAX\{\mathbf{f}_j|\mathbf{p}_j\in \mathbf{D}_i\}.
\end{equation}

\noindent
Then the aggregated features are registered to the corresponding block centroids. A python-style pseudo-code for an Exponential-based adaptive weight aggregator is shown in Algorithm~\ref{alg:expaw}.

\begin{figure*}[tbp]
\begin{center}
\includegraphics[width=0.98\linewidth]{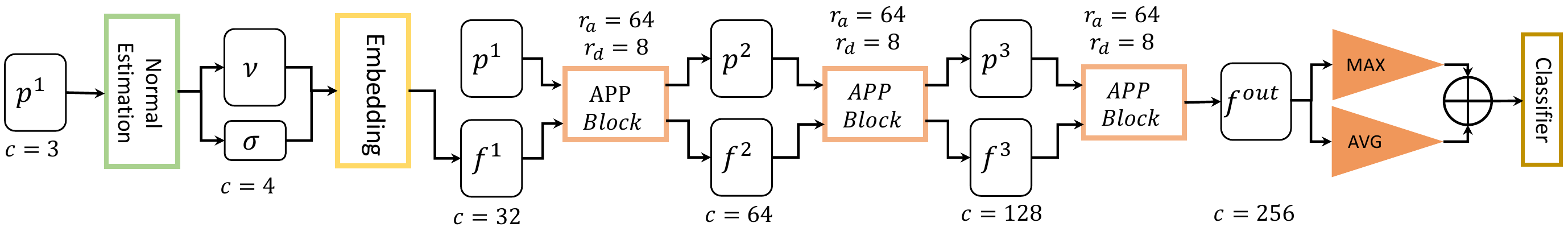}
\end{center}
% \vspace{-0.5cm}
   \caption{The whole structure of the proposed APP-Net. The channel of normal estimation is 3 or 4, which corresponds to whether to use the curvature.}
   \label{fig:a3pipeline}
\end{figure*}

\noindent
{\bf -The Full Structure of APP-Net}
This part introduces how to build an end-to-end network with the APP operator. As shown in Figure~\ref{fig:a3pipeline}, the input is the information of the point, including the position $xyz$, the normal and local curvature. An embedding layer consisting of [Linear($C_{in}\to32$), BatchNorm, LeakyReLU] lifts the input normal (and curvature) to an embedding with a dimension of 32. Three layers of the APP block are cascaded to aggregate neighbor features for every point. The APP operators and the block down sample are configured with the corresponding auxiliary rate $r_a=[64,64,64]$ and downsample rate $r_d=[8,8,8]$, respectively. A pooling layer (concatenates the results of AVG Pooling and MAX Pooling) outputs the global feature after the last APP layer. According to the global feature, a classifier consisting of MLPs predicts the probability for each category.

\subsection{Local Geometry from Online Normal and Curvature Estimation}
The normal and curvature reflect the local surface property. Previous methods often use the coordinates as input features and design a delicate and heavy local extractor to model the local geometric structure. In this work, we turn to directly feed the network with the explicit geometric descriptor, i.e., normal and curvature, to simplify the process of modeling geometric structure. Then the network can be more concentrated on learning the semantic relation. In point cloud, normal estimation is approximated by the problem of estimating the normal of a plane tangent to the surface~\cite{pcl}. In this paper, we adopt the simplest PCA-based method. A brief revision is present here to make the text more self-contained. For the centroid point $\mathbf{p}$, computing its local covariance matrix by

\begin{figure}[tbp]
\begin{center}
\subfigure[Ground Truth Normal.]{
% \label{fig:GT}
\includegraphics[width=0.35\linewidth]{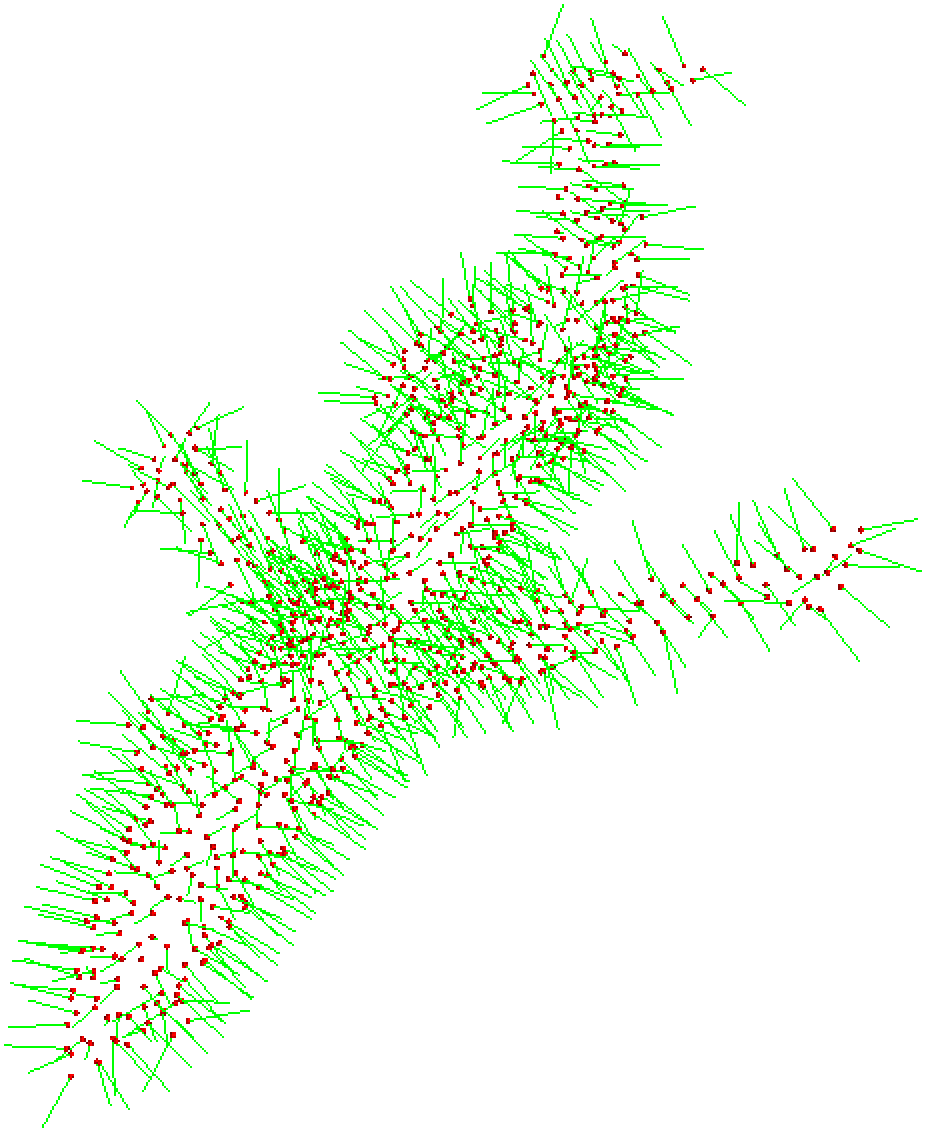}}
\subfigure[Estimated Normal.]{
% \label{fig:normalestimate}
\includegraphics[width=0.35\linewidth]{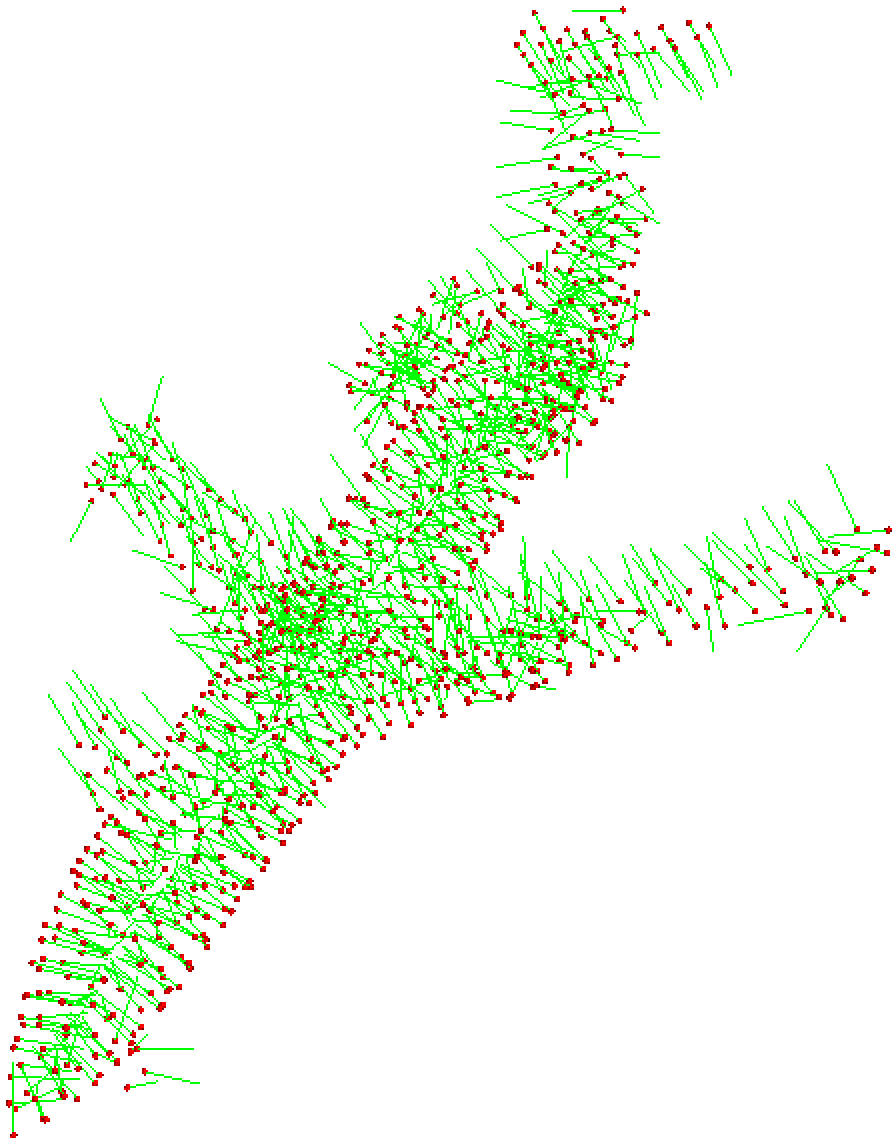}}
% \subfigure[Abs Normal.]{
% % \label{fig:absnormal}
% \includegraphics[width=0.25\linewidth]{fig/normal_abs.png}}
\end{center}
\vspace{-0.3cm}
   \caption{Different types of normals.}
\label{fig:normal}
\end{figure}

\begin{equation}
C = \frac{1}{|N(\mathbf{p})|}\sum_{\mathbf{p}_i\in N(\mathbf{p})}{(\mathbf{p}_i-\mathbf{p})\times(\mathbf{p}_i-\mathbf{p})^T},
\end{equation}

\begin{equation}
C\cdot \vec{v_{j}}=\lambda_j\cdot \vec{v_j}, j\in\{0,1,2\}
\end{equation}

\noindent
$\vec{v}_*$ and $\lambda_*$ represent the eigenvectors and eigenvalues of the covariance matrix, respectively. The eigenvector $\vec{v}$ corresponding to the minimum eigenvalue is the estimated normal. Supposed that $\lambda_0$ is the minimum eigenvalue, then the curvature $\sigma$ of the local surface is determined by 
\begin{equation}
\sigma = \frac{\lambda_0}{\lambda_0+\lambda_1+\lambda_2}
\end{equation}

%\noindent
A direction consistency check will flip those normal who do not orient towards a pre-specified viewpoint $(0,0,0)$ to alleviate the ambiguity in the normal direction. The comparison among different normals is depicted in Fig~\ref{fig:normal}.

\subsection{Discussion and Analysis on the APP-Net}
\subsubsection{Receptive Field Analysis}

\begin{figure}[tbp]
\begin{center}
\includegraphics[width=0.98\linewidth]{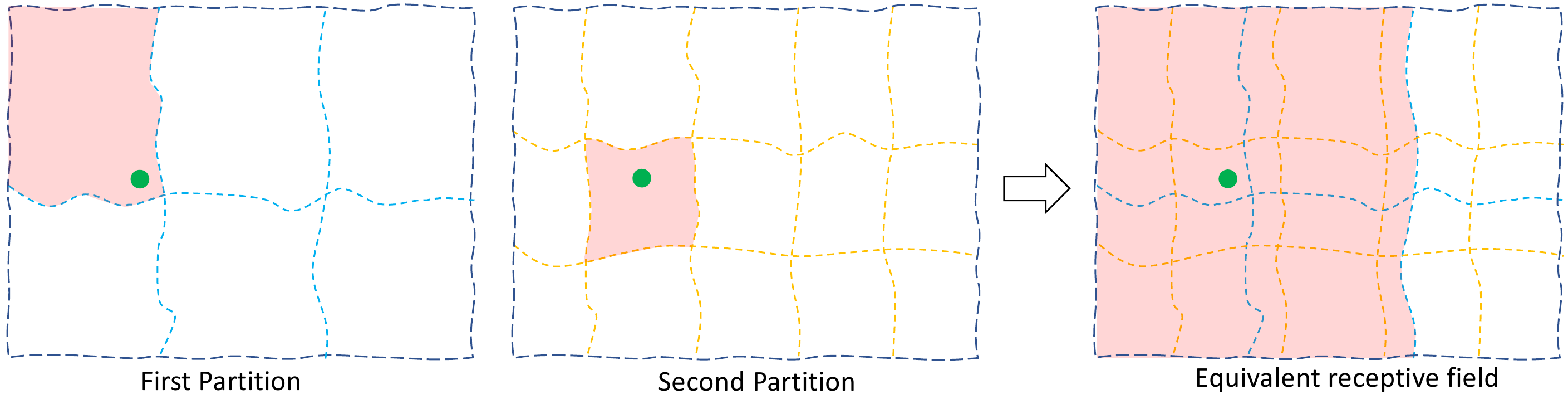}
\end{center}
\vspace{-0.5cm}
   \caption{Two-stage Block Partition. The first stage is in the push and pull step. The second stage is used for the block down sample. The whole point cloud are randomly partitioned into several non-overlapped blocks with different partition ratios. Thus the effect of the two-stage block partition is equivalent to the overlapped blocks.}
   \label{fig:receptivefield}
\end{figure}

The auxiliary point-based feature aggregation process includes two non-overlapped partitions, one for point mixing in the push step and pull step and the other for downsampling points. It's a common practice that large and expandable receptive fields are critical to learning good representation. As illustrated in Fig~\ref{fig:receptivefield}, the receptive field expands rapidly by combining the two non-overlapped partitions. A difference from the previous methods is that the expanded receptive field is irregular and random. Although introducing some uncertainties to the local descriptor, the random receptive field does not damage the global descriptor. And the global descriptor is more crucial to the classification task.

\subsubsection{Relation to Prior Work}

\noindent
{\bf Relation to PointNet} Both PointNet~\cite{PointNet} and the APP-Net's complexity are $\mathcal{O}(n)$ in each layer. However, the PointNet layer lacks point mixing, while the APP-Net introduces point mixing with the push and pull step, enabling information passing among points. Moreover, PointNet lacks a downsampling operation. Thus its high layers suffer intensive computations. In the APP-Net, the block-based down sample makes the high layers lightweight. 

\noindent
{\bf Relation to PointNet++ and its follow-up} PointNet++ and its follow-ups gather features from a sphere area around the center point, while the APP-Net gathers features from a random and irregular area. Another crucial difference is that PointNet++ imitates the convolution operation to accomplish receptive field expansion in one step, while the APP-Net does it in two stages. And the two-step style facilitates the linearization of feature extraction. One more difference is that we use the normal and curvature as the input to the network, while most of the previous work use the coordinates as the input.

\noindent
{\bf Relation to Transformer-like work} Both the local Transformer block and the APP block model the pair-wise relations in the local area. The self-attention in Transformer constructs a $N\times N$ relation matrix. However, in the APP block, we decompose the $N\times N$ relation matrix into a $N\times 1$ matrix and a $1\times N$ matrix, corresponding to the push step and pull step, respectively. This implies that the APP block is a linear version of the Transformer block and models a low-rank relation matrix.

\noindent
{\bf Relation to Conv-Deconv architecture} The APP block is similar to the Conv-Deconv operation pair, where the push step convolves the input from N points into M points, and the pull step deconvolves the M points back into N points. However, they are completely different designs. The convolution and deconvolution are independent steps and do not share the learnable parameters. And the intermediate M points are nonnegligible. However, in the APP block, the push and pull steps are highly coupled. They reuse the same learnable parameters, and the influence from the intermediate M points is reducible.

\subsubsection{Complexity Analysis} 
\label{sec:complexity}
We compare the overhead of the proposed auxiliary-based method with the point-wise-MLP layer. Given the input $\{P\in \mathbb{R}^{N\times 3}, F\in \mathbb{R}^{N\times C_{in}}\}$, we want to obtain a output of $\{P'\in \mathbb{R}^{M\times 3}, F'\in \mathbb{R}^{M\times C_{out}} \}$. Table~\ref{tab:appcomplexity} and Table~\ref{tab:pn2complexity} show the computation complexity and memory consumption for the APP block and a single scale single MLP PointNet++ block, respectively. $K$ denotes the number of nearest neighbors. The dominating computation is $2*N\times C_{in}\times C_{out}$ for APP block, $M\times K\times C_{in}\times C_{out}$ for the PointNet++ block. Due to $M\times K \gg 2\times N$ in practice, the APP block owns lower computations. In the stable implementation of PointNet++~\cite{PointNet2PyTorch}, $M\times K=16\times N$, which induces $8\times$ computations. As to the memory, the dominating part is $3\times N\times C_{in}+N\times C_{out}$ for APP block, $M\times K\times C_{in}+M\times K\times C_{out}$ for PointNet++. Similarly, the APP block owns several times lower memory consumption. Meanwhile, the above analyses are based on the single scale and single MLP settings. If employing the commonly used multiple scales or multiple MLP for the PointNet++, the overhead advantages for APP block would be more obvious.

\begin{table}[t]
	\caption{APP block overhead analysis.}
	\label{tab:appcomplexity}
\begin{center}
\begin{tabular}{c|c|c}
\hline
      \makecell[c]{Step}&\makecell[c] {Computation} & Memory  \\ \hline
\makecell[c]{Position Encoding} & $N\times 3\times C_{in}$      & $N\times C_{in}$\\ \hline
Push                            &\makecell[c]{$N\times C_{in}$ \\ $\frac{N}{r_a}\times C_{in}$}&\makecell[c]{$N\times C_{in}$ \\ $\frac{N}{r_a}\times C_{in}$} \\ \hline
Pull                            &$N\times C_{in}$       &$N\times C_{in}$ \\ \hline
Channel Mixing &$N\times C_{in}\times C_{out}$  &$N\times C_{out}$ \\ \hline
Block Pool &-   & $M\times C_{out}$ \\ \hline
\end{tabular}
\end{center}
\end{table}

\begin{table}[t]
	\caption{PointNet++ block overhead analysis.}
	\label{tab:pn2complexity}
\begin{center}
\begin{tabular}{c|c|c}
\hline
      \makecell[c]{Step}&\makecell[c] {Computation} & Memory  \\ \hline
Group & -      & $M\times K\times (C_{in}+3)$\\ \hline
MLP                            &$M\times K\times(C_{in}+3)\times C_{out}$& $M\times K\times C_{out}$ \\ \hline
Pooling                            & -       &$M\times C_{out}$ \\ \hline

\end{tabular}
\end{center}
\end{table}

\section{Experiments}
To explore the characteristics of our method, we conduct extensive experiments on a synthetic dataset and a real scene dataset. Our method has achieved competent accuracy while maintaining very low overhead and high efficiency compared with existing methods. We further conduct some ablation studies on Section~\ref{sec:ablation} to test how every module works.
 
\subsection{Settings}

Our experiments are evaluated on a server with one Tesla V100 GPU. Most of the projects are implemented with PyTorch~\cite{paszke2019pytorch}. For all the tasks, we use the default Adam~\cite{kingma2014adam} to update the network parameters. And we use the cosine learning rate scheduler~\cite{cosannel} to update the learning rate, with an initial learning rate of 2e-3. The minimum learning rate threshold is set to 2e-4. The cycle for the cosine scheduler is $T_{max}=200$. For all the experiments, we train the network for 300 epochs with a training batch size of 32 and a test batch size of 16; we use the first epoch to warm up the training process. All the results of the comparison methods are obtained in three ways: 1. careful reproduction in our environment to prevent the unfairness caused by machine development (For fairness, if we fail to reproduce the public results, we will adopt the following two ways); 2. the reported results in the original papers; 3. the updated results on the public websites or other published papers. 

For simplicity, we denote each variant of APP-Net with the combination of "basic operator+aggregator type" in the following part. The basic operators contain \{Exp, Sin, Cos\} and the aggregator type contains \{AW, PW\}. 'AW' means to use the adaptive weight style, and 'PW' denotes the point-wise MLP style.

\subsection{Shape Classification on ModelNet40}

\begin{table*}[t]
\begin{center}
	\caption{Classification on ModelNet40. We report the overall accuracy, train speed, test speed, and the number of parameters of some baselines. 5k denotes 4096 points, and the 7k for KPConv means using around 7,000 points. 'P' and 'N' means using point and ground truth normal, respectively. {\bf Bold} number denotes the best one.}\label{tab:mn40}
\begin{tabular}{c|c|c|c|c|c}
\hline
Model      & Inputs  &\makecell[c]{Train Speed\\(samples/s)}  & \makecell[c]{Test Speed\\(samples/s)}  & Param. & OA(\%)\\ \hline
PointNet~\cite{PointNet}&1k P&960.9&1422.4&3.5M&89.2\\
Pointnet++~\cite{PointNet++}&1k &352.2&730.2&1.41M&90.7\\
PointNet++~\cite{PointNet++}&5k P+N&-&-&1.41M&91.9\\
PointCNN~\cite{pointcnn}&1k P&-&-&-&92.5\\
PointConv~\cite{pointconv} &1k P+N&104.5&76.4&18.6M&92.5\\
KPConv~\cite{KPConv}&7k P&211.7&717.7&15.2M& 92.9\\
DGCNN~\cite{DGCNN}&1k P&-&-&-&92.9\\
RS-CNN~\cite{liu2019relation}&1k P&-&-&-&92.9\\
DensePoint~\cite{liu2019densepoint}&1k P&-&-&-&92.8\\
ShellNet~\cite{shellnet}&1k P&551.3&1077.5&{\bf0.48M}&93.1\\
PointASNL~\cite{PointASNL}&1k P+N&285.9&316.4&3.2M&93.2\\
PosPool~\cite{closerlook3d}&5k P&51.0&59.5&18.5M&93.2\\
Point Trans.~\cite{engel2021point} &1k P&-&-&-&92.8\\
GBNet~\cite{qiu2021geometric}& 1k P&17.7&175.9&8.39M&{93.8}\\
GDANet~\cite{xu2021learning}& 1k P&29.8&273.3&0.93M&93.4\\
PA-DGC~\cite{xu2021paconv}&1k P&-&-&-&93.6\\
MLMSPT~\cite{han2021point}&1k P&-&-&-&92.9\\
PCT~\cite{guo2021pct}&1k P&115.7&1044.9&2.88M&93.2\\
Point Trans.~\cite{zhao2021point}&1k P&67.1&82.3&12.9M&93.7\\
CurveNet~\cite{muzahid2020curvenet}&1k P&89.9&112.9&2.04M&{93.8}\\
PointMLP~\cite{ma2022rethinking}&1k P&60.4&169.0&12.6M&{\bf94.1}\\
PointMLP-elite~\cite{ma2022rethinking}&1k P&240.1&632.8&{0.68M}&93.6\\ \hline
APP-Net(Exp+AW)& 5k P&{785.9}&{1451.8}&0.77M&93.0\\ \hline
APP-Net(Exp+AW) & 5k P+N&{\bf1440.6}&{\bf 2155.5}&{0.77M}& 94.0    \\
APP-Net(Cos+AW) & 5k P+N&1174.1&1971.2&0.79M& 93.5    \\
APP-Net(Cos+PW) & 5k P+N&1274.2&2107.6&0.77M& 93.4    \\
APP-Net(Sin+AW) & 5k P+N&1224.3&1995.1&0.79M& 93.2    \\
APP-Net(Sin+PW) & 5k P+N&1264.3&2095.1&0.77M& 93.4    \\ 
\hline

\end{tabular}
\end{center}
	%\vspace{-0.42cm}
\end{table*}

\begin{table*}[t]
	\caption{Time cost analysis for each module of the APP-Net.}\label{tab:timeanalysis}
\begin{center}
\begin{tabular}{c|c|c|c|c|c|c}
\hline
     & \makecell[c]{Normal\\ Estimation}&\makecell[c] {Feature \\ Embedding} & Layer 1 & Layer 2 & Layer 3 & Classifier \\ \hline
APP-Net   & 2$ms$     & 0.13$ms$              &     1.2$ms$                       & 1.4$ms$             & 1.1$ms$   & 0.07$ms$    \\ \hline
PointNet++~\cite{PointNet++}&-&-&20.01$ms$&12.8$ms$&0.88$ms$&0.27$ms$\\
\hline
\end{tabular}
\end{center}
\end{table*}

ModelNet40~\cite{wu20153d} is the most influential benchmark for 3D classification task, consisting of 40 common categories. The point-cloud-type data is synthesized by randomly sampling the surface of 3D CAD models, with a training set of 9,843 samples and a testing set of 2,468 samples. We report the most widely used metric, Overall Accuracy, on the testing set. For fairness, we do not adopt the voting strategy for all the methods(which often improves the accuracy by about 0.4\% for some methods). Besides, we also report the computation overhead. As shown in Table~\ref{tab:mn40}, we achieve comparable accuracy to these SOTAs and maintain a very efficient running speed. The speed is measured by $\frac{Total \ Samples}{Total\ Inference\ Time}$. Most of the baseline methods obtain a faster speed than the one reported in the previously published papers. We believe it is mainly attributed to the machine difference and the adoption of the optimized CUDA operations. Using 1024 points is a standard setting for this task. There are also some methods choosing to input more (4096 or more) points to boost the result at the price of a heavier burden. For ModelNet40, the proposed APP-Net is fed with 4096 points while still running faster than all the other baselines of 1024 points. And it's {\bf $3\times$} faster than the PointNet++~\cite{PointNet++}, {\bf $19\times$} faster than CurveNet~\cite{muzahid2020curvenet} during test, which is coherent with the analysis in section~\ref{sec:complexity}. APP-Net has a clear speed advantage over the other methods even with the online estimated normal.

\subsection{Shape Classification on ScanObjectNN}

Considering the saturation of ModelNet40~\cite{wu20153d} and challenging real-world cases, Uy et al. propose the ScanObjectNN~\cite{uy2019revisiting}, collecting by scanning the indoor objects. The real-world data often face some annoying issues, like the cluttered or occluded by fore/background. So ScanObjectNN reveals the great potential to promote the classification application in the real world.

%\begin{wraptable}[25]{r}{1.0\linewidth}
\begin{table*}[t]%{r}{1.0\linewidth}
 \begin{center}
 %\vspace{-0.9cm}
	\caption{Classification on ScanObjectNN. The input for APP-Net contains 1024 points. We run the experiment five times and report the mean$\pm$std. {\bf Bold} number denotes the best one. *$\ddag$ uses more learnable parameters (still less than most of the baselines). }\label{tab:sonn}
\vspace{0.2cm}
\begin{tabular}{c|c|c|cc}
\hline
Methods      & Inputs  & OA(\%) &\makecell[c]{Train Speed\\(samples/s)} &\makecell[c]{Test Speed\\(samples/s)}\\ \hline
PointNet~\cite{PointNet}   & Point                      & 68.2  &960.9& 1422.4 \\
SpiderCNN~\cite{xu2018spidercnn} &Point&73.7&-&-\\
PointNet++~\cite{PointNet++} &Point&77.9&352.2&730.2\\
DGCNN~\cite{DGCNN} &Point& 78.1&-&-\\
PointCNN~\cite{pointcnn}&Point&78.5&-&-\\
BGA-DGCNN~\cite{uy2019revisiting}&Point&79.7&-&-\\
BGA-PN++~\cite{uy2019revisiting}&Point&80.2&-&-\\
DRNet~\cite{qiu2021dense}&Point&80.3&-&-\\
GBNet~\cite{qiu2021geometric}&Point&80.5&-&-\\
SimpleView~\cite{goyal2021revisiting}&Multi-view&$80.5\pm0.3$&-&-\\
PRANet~\cite{cheng2021net}&Point&82.1&-&-\\
MVTN~\cite{hamdi2021mvtn}&Multi-view&82.8&-&-\\
PointMLP~\cite{ma2022rethinking}&Point&$85.4\pm0.3$&60.4&169.0\\
PointMLP-elite~\cite{ma2022rethinking}&Point&$83.8\pm0.6$&240.1&632.8\\
\hline
APP-Net(Exp+AW)&Point&$84.3\pm0.3$&{\bf1633.0}&2442.0\\
APP-Net(Cos+AW)&Point&$84.2\pm0.1$&1377.3&2343.1\\
APP-Net(Cos+PW)&Point&$84.4\pm0.2$&1405.4&2395.7\\
APP-Net(Sin+AW)&Point&$84.7\pm0.1$&1394.2&2387.7\\
APP-Net(Sin+PW)&Point&$84.4\pm0.1$&1359.4&{\bf2471.7}\\
APP-Net(Cos+AW){$\ddag$}&Point&$86.1\pm0.2$&1239.7&2069.9\\
APP-Net(Sin+AW){$\ddag$}&Point&${\bf87.0\pm0.2}$&1153.9&2023.9\\
\hline
\end{tabular}
\end{center}
\end{table*}
%\end{wraptable}

%\vspace{-0.2cm}

According to the common practice in other work, we use the hardest variant PB\_T50\_RS to conduct our experiments. The whole dataset contains a training set with 11416 samples and a testing set with 2882 belonging to 15 categories. We choose the most representative point-based and multi-view methods as the baselines. The normal we put into the network is computed online, and the duration of normal estimation is considered in the speed test. The overall accuracy is shown in Table~\ref{tab:sonn}. Following SimpleView~\cite{goyal2021revisiting}, we report the mean$\pm$std. We outperform all the baseline methods, including the PointMLP~\cite{ma2022rethinking}. The lightweight "Exp+AW" version is $14\times$ faster than PointMLP~\cite{ma2022rethinking} and $3.8\times$ faster than the lightweight PointMLP-elite. With more parameters, the APP-Net achieves the best accuracy among all the methods. And the large version is still faster than all the other methods. In Table~\ref{tab:timeanalysis}, we report the time cost of each module under test mode, with batch size=16 and $N$ =1024. Each layer only costs $1.1\sim1.4$ms. %Although the normal estimation occupies a relatively long time(2ms), it's worthwhile because the estimated normal promotes our simple architecture to achieve better accuracy.

\subsection{Ablation Studies}
\label{sec:ablation}

There are some fine-designed structures in APP. To test their functionality and substitutability, we conducted some ablation studies and analyses based on the ScanObjectNN.

\noindent
{\bf The Advantages of the Reducible Operation}
One of the core designs of the APP is the reducible operation pair: Push-Step and Pull-step. They make each point's representation independent of the auxiliary point. To verify the effectiveness, we design two types of non-reducible operations: 1. use two different mapping functions to compute the spatial relation for the Push-step and Pull-step, respectively; 2. add non-linear operation, i.e., BatchNorm and Leaky ReLU, to the spatial relation. The results in Table~\ref{tab:reducible} show that the reducible operation is non-trivially superior to the others in accuracy.
\begin{table}[t]
\begin{center}
	\caption{The reducible operation. We explore whether the reducible attribute is necessary and a new instantiation of the reducible philosophy. The best one is colored with {\bf Bold}. Tested with the "Exp+AW" aggregator.}\label{tab:reducible}
% 	\vspace{-0.2cm}
\begin{tabular}{c|c}
\hline
     & \makecell[c]{OA(\%)} \\ \hline
Reducible   & {\bf84.3} \\ \hline
\makecell[c]{Not Reducible [Different Mapping Function]}   & 82.8 \\ \hline
\makecell[c]{Not Reducible [Non-linear Mapping Function]}   & 83.2 \\ \hline
% \hline
\end{tabular}
\end{center}
\end{table}

\begin{figure}[!t]
    \centering
    \includegraphics[width=0.98\linewidth]{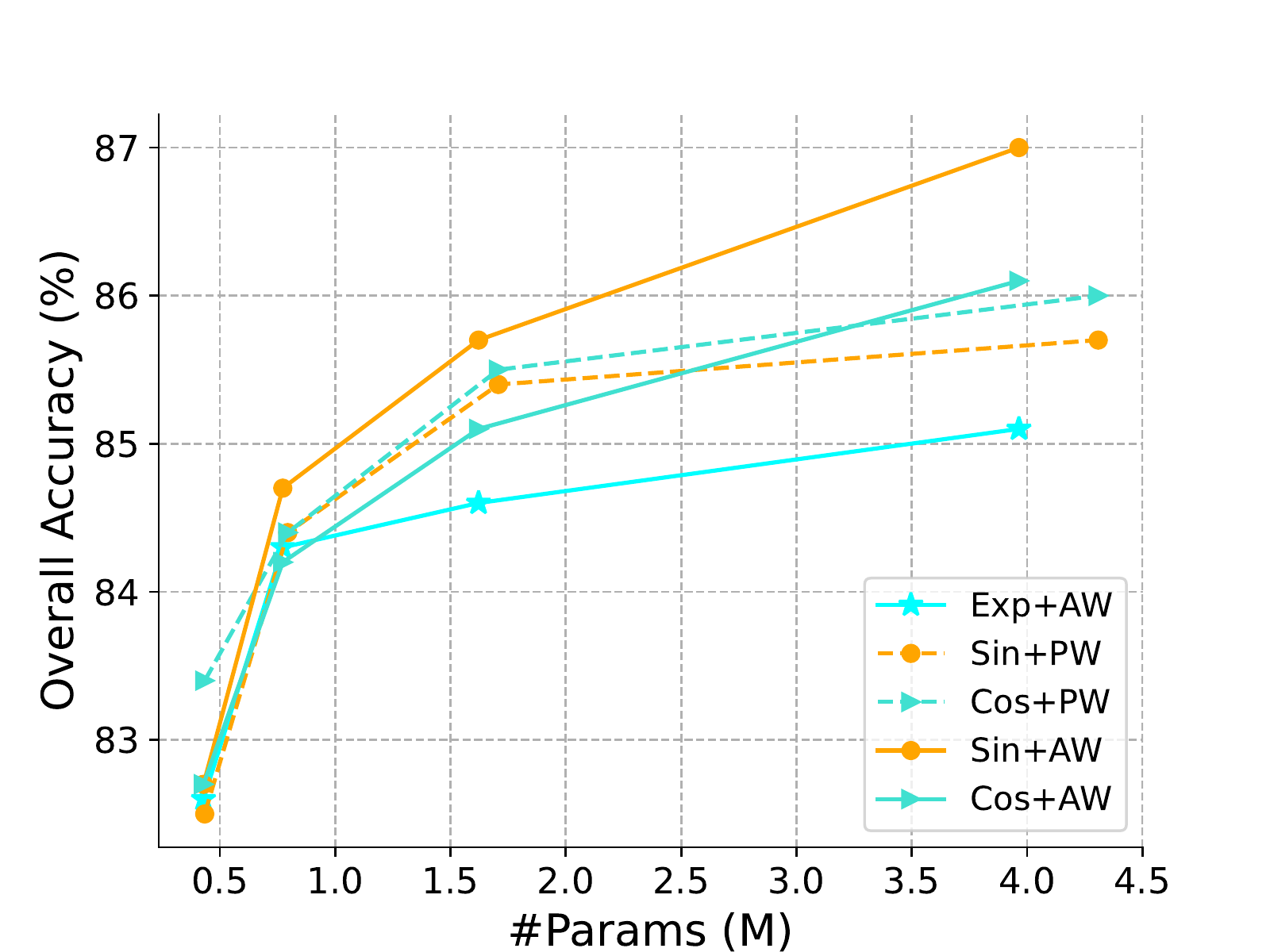}
    \caption{Performance gains with the growing model size.}
    \label{fig:scalability}
\end{figure}

\noindent
{\bf The Input to the Network}
Unlike previous work, which takes the original coordinates as the input, we use the estimated normal and curvature as explicit descriptors to represent the local geometry. Table~\ref{tab:input} shows the comparisons among different inputs to different configured networks. The results clearly indicate that the estimated normal and curvature significantly improve the performance. Meanwhile, when fed with the $xyz$, the APP-Net achieves PointNet++'s~\cite{PointNet++} and DGCNN's~\cite{DGCNN} performance with a clear overhead advantage.

\begin{table}[t]
\begin{center}
	\caption{Different input to the network.}\label{tab:input}
\begin{tabular}{c|c|c|c|c}
\hline
     &Sin+AW& Cos+AW&Sin+PW&Cos+PW\\ \hline
$xyz$ & 76.1 & 78.2 & 79.8 &  79.4\\ \hline
$normal$ & 83.8 & 83.0 & 83.3 &  83.7\\ \hline
\makecell[c]{$normal$+\\$curvature$} & {\bf84.7} & {\bf84.2} & {\bf84.4} &  {\bf84.4}\\
\hline
\end{tabular}
\end{center}
\end{table}

\noindent
{\bf Model Scalability}
In this part, we scale up the network by enlarging the number of feature channels to test the scalability of the APP-Net. According to Figure~\ref{fig:scalability}, the performance grows along with the number of channels. The Sin-based and Cos-based networks outperform the Exponential Function; we think its due to the value range of the Exponential Function being unbounded and growing rapidly, thus making it hard to optimize. In the point-wise manner, the Cos-based method is marginally better than the Sin-based method, while in the adaptive weighting variant, the relation are reversed.

\begin{table}[t]
\begin{center}
	\caption{Combinations of the APP layer and PointNet++ layer. 'A' denotes using APP layer and 'P' denotes using PointNet++ layer. Tested with the "Exp+AW" aggregator.}\label{tab:combination}
\begin{tabular}{c|c|c|c|c|c}
\hline
 Layer 0 & Layer 1& Layer 2 & \makecell[c]{OA\\(\%)}  & \makecell[c]{Speed\\(samples/s)} &\makecell[c]{Memory\\(MB)} \\ \hline
A&A&A &{\bf84.3}&2442.0&1393\\ \hline
A&A &P&83.7&2058.6&3361\\ \hline
A&P&P&83.7&1762.4&5535\\ \hline
P&P&P&84.1&1353.3&5535\\ \hline
\end{tabular}
\end{center}
\end{table}

\noindent
{\bf Co-operation with other operators}
According to Table~\ref{tab:timeanalysis}, the first layer in PointNet++ occupies a large ratio of time. We layer-wisely replace the APP layer in our network with the PointNet++ layer to explore the effect of combining different layers. According to Table~\ref{tab:combination}, the overall accuracy of every combination is comparable, while replacing the PointNet++ layer with the APP layer greatly accelerates the network and reduces memory consumption. Meanwhile, the results of the 3 layers PointNet++ is more effective and efficient than the one in Table~\ref{tab:sonn}. We believe it is because we implement it with the single MLP layer and the input of the network is the normal and curvature rather than the coordinates. The identical memory consumption in "A+P+P" and "P+P+P" settings are caused by the internal memory allocation mechanism of PyTorch.

\noindent
{\bf Farthest Point Sample versus Random Sample}
We use a random sample in the blockdown sample module. According to Table~\ref{tab:downsample}, although the farthest point sample produces a more uniform subset of the input point cloud, it does not lead to better accuracy. And it is slightly slower than the random sample in our experiments.

\begin{table}[t]
	\centering
	\caption{Down Sample Methods. 'RS' denotes random sample and 'FPS' denotes farthest point sample.}
	\begin{tabular}{c|c|c|c|c}
		\hline
		\multirow{2}{*}{Configs}    & \multicolumn{2}{c|}{OA(\%)} &  \multicolumn{2}{c}{\makecell[c]{Speed\\(samples/s)}}           \\ \cline{2-5} 
		& Exp+AW        & Sin+AW    & Exp+AW        & Sin+AW  \\ \hline
		\multicolumn{1}{c|}{RS} & {\bf84.3}          & {\bf84.7}             & {\bf2442.0}  &     {\bf2387.7}                   \\ \hline
		\multicolumn{1}{c|}{FPS} & 83.8          & 84.1             & 
		2320.4&2235.2
		                        \\ \hline
	\end{tabular}\label{tab:downsample}
\end{table}

\noindent
{\bf Whether to use feature difference in the Point-wise MLP aggregator}
To eliminate the calculation for the point-wise MLP style aggregator, we reuse the concatenation of feature and position encoding for the push and pull steps. This induces the aggregator to model the feature difference. To explore the effect of modeling the feature difference and the original feature, we decouple the calculation of the push and pull steps. According to Table~\ref{tab:pwdiff}, the modeling of the original feature marginally hits a better accuracy. However, it slows down the network. Taking the results in Table~\ref{tab:sonn} into account, the speed around 2000 samples/s can be also met through enlarging the channels, and the enlarged version achieved even better accuracy. To keep the architecture simple, we model the feature difference as the standard version.

\begin{table}[t]
	\centering
	\caption{Pointwise MLP style analysis}
	\begin{tabular}{c|c|c|c|c}
		\hline
		\multirow{2}{*}{Pulls}    & \multicolumn{2}{c|}{OA(\%)} &  \multicolumn{2}{c}{\makecell[c]{Speed\\(samples/s)}}           \\ \cline{2-5} 
		& \makecell[c]{Sin\\+PW}        & \makecell[c]{Cos\\+PW}    & \makecell[c]{Sin\\+PW}        & \makecell[c]{Cos\\+PW}  \\ \hline
		\multicolumn{1}{c|}{\makecell[c]{$W\times[f_i, \phi(A(p_i)-\phi(p_i))]$}} & 84.4          & 84.4             & {\bf2471.7}  &     {\bf2395.7}                   \\ \hline
		\multicolumn{1}{c|}{$W\times[0, \phi(A(p_i))-\phi(p_i)]$} & {\bf84.7}          & {\bf85.0}             & 
		2143.2&1992.0
		                        \\ \hline
	\end{tabular}\label{tab:pwdiff}
\end{table}

\begin{table}[!t]
\begin{center}
	\caption{Different ways to model the relations among neighbors. Tested with the "Exp+AW" aggregator.}\label{tab:posenc}
	\vspace{-0.2cm}
\begin{tabular}{c|c}
\hline
   Configs  & \makecell[c]{OA(\%)} \\ \hline
\makecell[c]{Global Position Encoding\\$\phi(\mathbf{p}_i)$}    & {\bf84.3} \\ \hline
\makecell[c]{Local Position Encoding\\$\phi(\mathbf{p}_i-A(\mathbf{p}_i))$}   & 82.8 \\ \hline
\makecell[c]{No Position Encoding\\$\mathbf{p}_i$}   & 82.5 \\ \hline
\makecell[c]{Feat+Global Position Encoding\\$\mathbf{f}_i+\phi(\mathbf{p}_i)$}   & 83.6 \\ \hline
\makecell[c]{Concat[Feat, Global Position] Encoding\\$[\mathbf{f}_i,\phi(\mathbf{p}_i)]$}   & 82.1 \\
\hline
\end{tabular}
\end{center}
\end{table}

\noindent
{\bf Different Relation Modeling Methods} 
We compare different ways of encoding points' relations in this part. The variants include using the global position encoding, local position encoding, or directly computing the spatial relation without using position encoding. The local position is computed by subtracting the corresponding center point. Considering that the transformer measures the relation between features, we also try to combine the feature relation in the APP. Results in Table~\ref{tab:posenc} show that the global position encoding obtains better results than the local one. We think it is due to the instability of the local position caused by the random block partition, which hinders the network convergence during training. Besides, the feature relation does not provide a positive effect. We guess it wrongly builds the dependencies between the position and feature. This also explains why the "Feat+Global Position Encoding" performs better than "Concat[Feat, Global Position] Encoding" since the former decouples the process of learning feature relation and spatial relation.

\noindent
{\bf Feature Updating Style} At the end of the pull step, every point's feature is updated by concatenating the output with the original feature and sending it to a $\delta(*)$ function. Among the comparisons, we try to remove some parts of it or leverage a residual structure. As shown in Table~\ref{tab:sigma}, The concatenation manner is superior to the other configurations. Moreover, in the last two rows, we remove the $\delta(*)$ function or the whole APP block; the results clearly state their indispensability.

\begin{table}[t]
\begin{center}
	\caption{Different ways to update the feature after the Pull-step. Tested with the "Exp+AW" aggregator.}\label{tab:sigma}
\begin{tabular}{c|c|c|c|c}
\hline
     &\makecell[c]{Concat\\$\mathbf{g}_i=\delta([\mathbf{g}_i, \mathbf{f}_i])$} & \makecell[c]{Not Concat\\$\mathbf{g}_i=\delta(\mathbf{g}_i)$}   &\makecell[c]{Res Feature\\$\mathbf{g}_i=\delta(\mathbf{g}_i)+\mathbf{f}_i$}   & \makecell[c]{Identity\\$\mathbf{g}_i = \mathbf{g}_i$}   \\ \hline
\makecell[c]{OA\\(\%)}  & \bf{84.3} & 77.8 & 83.7 &  79.3\\
\hline
\end{tabular}
\end{center}
\end{table}

\noindent
{\bf Network Depth}
We explore how the network depth affects the performance. In the 2-layer and 3-layer version, we adopt the same $r_a$ and $r_d$. In the 4-layer version, due to the original number of the point being 1024, after two downsample operations, the remaining points are insufficient to support a large $r_a$, so we adopt a smaller $r_a=16$ in the last two layers. According to Table~\ref{tab:depth}, the 3-layer version achieves the best performance. 

\begin{table}[t]
\begin{center}
	\caption{The influence of the network depth. With different layers, the rate will be adjusted adaptively.}\label{tab:depth}
\begin{tabular}{c|c}
\hline
  Configs   & \makecell[c]{OA(\%)} \\ \hline
\makecell[c]{2 layers, $r_d$=[8,8], $r_a$=[64,64]}   & 83.0 \\ \hline
\makecell[c]{3 layers, $r_d$=[8,8,8], $r_a$=[64,64,64]}   & {\bf 84.3} \\ \hline
\makecell[c]{4 layers, $r_d$=[4,4,8,8], $r_a$=[64,64,16,16]}   & 83.9 \\ 
\hline
\end{tabular}
\end{center}
\end{table}

\begin{table}[t]
\centering
	\caption{Different rate configurations for the APP-Net.}\label{tab:rate}
\begin{tabular}{c|c}
\hline
    Rate & \makecell[c]{OA(\%)} \\ \hline
\makecell[c]{$r_d$=[8,8,8],$r_a$=[64,64,64]}   & {\bf84.3} \\ \hline
\makecell[c]{$r_d$=[4,4,4],$r_a$=[64,64,64]}   & 84.1 \\ \hline
\makecell[c]{$r_d$=[16,16,16],$r_a$=[64,64,64]}   & 83.5 \\ \hline
\makecell[c]{$r_d$=[8,8,8],$r_a$=[32,32,32]}   & 82.6 \\ \hline
\makecell[c]{$r_d$=[8,8,8],$r_a$=[32,64,64]}   & 83.7 \\ \hline
\makecell[c]{$r_d$=[8,8,8],$r_a$=[64,64,32]}   & 83.1 \\
\hline
\end{tabular}
\end{table}

\begin{table}[!t]
\centering
	\caption{Different pooling policies.}\label{tab:pooling2}
\begin{tabular}{c|c}
\hline
  Configs   & \makecell[c]{OA(\%)} \\ \hline
AVG+MAX Pooling   & {\bf84.3} \\ \hline
MAX Pooling   & 82.1 \\ \hline
AVG Pooling   & 79.7 \\ \hline
Position Adaptive Pooling  & 79.9 \\ 
\hline
\end{tabular}
\end{table}

\noindent
{\bf Down Sample Rate}
The rate $r_a$ and $r_d$ serve as the network's kernel size. They control the receptive field for each center point. For simplicity, we adopt the same $r_d$ for all layers in each variant. Results in Table~\ref{tab:rate} imply that a large $r_a$ for the auxiliary point generation is critical to a better representation, especially at the high level (according to the last two rows). And the result is relatively not sensitive to the rate $r_d$ for aggregation.

\noindent
{\bf Pooling Policies in the Aggregation}
We have tested with the common pooling policies to explore a proper aggregation operation for the second block partition. The position-adaptive pooling aggregates local points weighted by the reciprocal distance in Euclidean space. The results in Table~\ref{tab:pooling2} show that combining the local mean context and the most representative feature can achieve better performance for the classification task.

\subsection{Limitations}

We note the following limitations of this work:

\begin{enumerate}
    \item The proposed operator is designed for the classification task. However, this paper does not explore its generalization to other dense estimation tasks, like segmentation. Although it does not harm the global descriptor,  the random receptive field may introduce noise to the local descriptor. So the generalization to other tasks is challenging.
    \item In the synthetic dataset, the ground truth normal is necessary for the APP-Net to classify some hard cases (confusing with similar categories). This is because the geometry information is too dependent on the estimated noisy normal and curvature. A better estimation method or a lightweight geometric learning layer may alleviate it.

\end{enumerate}

\section{Conclusions And Future Work}
% \vspace{-0.2cm}
This paper proposes a novel APP operator aggregating local features through auxiliary points. It avoids redundant memory consumption and re-computation of the source point. Furthermore, the auxiliary points' influence is reducible, allowing the method to preserve more details. Experiments on the synthetic and real-scene datasets show a good balance between performance, speed, and memory consumption in the classification task. Especially in speed, it outperforms all the previous methods significantly. Our future goal is to extend this method to more tasks, like semantic segmentation and object detection.

% references section
\bibliographystyle{IEEEtran}
\bibliography{egbib.bib}

\begin{comment}
\bf{If you include a photo:}\vspace{-33pt}
\begin{IEEEbiography}[{\includegraphics[width=1in,height=1.25in,clip,keepaspectratio]{fig1}}]{Michael Shell}
Use $\backslash${\tt{begin\{IEEEbiography\}}} and then for the 1st argument use $\backslash${\tt{includegraphics}} to declare and link the author photo.
Use the author name as the 3rd argument followed by the biography text.
\end{IEEEbiography}

\vspace{11pt}

\bf{If you will not include a photo:}\vspace{-33pt}
\begin{IEEEbiographynophoto}{John Doe}
Use $\backslash${\tt{begin\{IEEEbiographynophoto\}}} and the author name as the argument followed by the biography text.
\end{IEEEbiographynophoto}

\end{comment}

\vfill

\end{document}